# A Cognition-Emotion-Personality Framework for Modeling Human-Like Awareness and Behavior in Emergency Evacuations

**Zoi Lygizou**[1,2*], **Michalis Zervas**[1], **Helena G. Theodoropoulou**[1], **Vasilis Zafeiropoulos**[1], **Dimitris Kalles**[2], **Chairi Kiourt**[1]

[1] Athena Research Centre, Xanthi, Greece

[2] Hellenic Open University, Patras, Greece

[*]Corresponding author: Zoi Lygizou (zoi.lygizou@ac.eap.gr)

**Emails**:
*zoi.lygizou@ac.eap.gr; michalis.zervas@athenarc.gr; hetheod@athenarc.gr; v.zafeiropoulos@athenarc.gr; kalles@eap.gr; chairiq@athenarc.gr*

## Abstract

Agent-based evacuation simulations are widely used to analyze crowd behavior during emergency situations; however, many existing models rely on simplifying assumptions such as perfect event awareness, complete exit knowledge, and fully rational decision-making. These assumptions limit their ability to reproduce the complex, heterogeneous, and often non-optimal behaviors observed in real evacuations. This paper proposes an extended agent-based evacuation framework that integrates cognitive, emotional, social, and personality-driven mechanisms into a unified model of human behavior under uncertainty. In particular, the framework incorporates (i) a dynamic event awareness mechanism based on a continuous Event Certainty Level, (ii) a memory-based representation of exit knowledge subject to acquisition, forgetting, and recall, (iii) a continuous fear level in which panic emerges as a high-intensity state influencing perception, decision-making, movement, and physical interactions, and (iv) a personality representation based on the OCEAN framework, in which each agent is characterized by a vector of personality traits. The Neuroticism dimension is explicitly integrated into the fear system. Rather than introducing new behavioral rules, Neuroticism acts as a modulation factor that systematically affects multiple components of emotional dynamics, including fear generation, fear escalation over time, social contagion, and recovery from panic. Heterogeneous evacuation behavior is further captured through individualized decision thresholds, allowing agents to differ in their sensitivity to perceived risk. The proposed framework is implemented in a controlled simulation environment and evaluated through a series of experiments designed to isolate the effects of spatial familiarity, memory robustness, decision sensitivity, emotional dynamics, and personality-driven variability on evacuation outcomes. The results demonstrate that incorporating cognitive, emotional, and personality-related processes significantly alters evacuation dynamics, leading to reduced evacuation efficiency and the emergence of realistic crowd phenomena such as delays, confusion, injuries, falls, trampling, and socially driven behaviors. In particular, the presence of fear and panic is shown to play a critical role in shaping both individual behavior and collective outcomes. Overall, the proposed framework provides a more realistic representation of human behavior in emergency evacuations and enables systematic investigation of the interaction between cognition, emotion, personality and crowd dynamics.

# 1 Introduction

Agent-based evacuation simulations are widely used to study crowd behavior during emergency situations. Traditional models often assume that all individuals instantly perceive an emergency event, such as a fire or explosion, and begin evacuation simultaneously [1]. In such simulations, agents are typically fully aware of the locations of all exits, immediately decide to evacuate, and select the nearest exit in an organized manner. Under these assumptions, panic-induced trampling, injuries, and fatalities do not occur, and all agents successfully evacuate the environment.

However, this idealized scenario rarely reflects real-world conditions [1]. Awareness of an ongoing incident usually emerges gradually and spreads unevenly throughout a crowd. Accurately modeling how knowledge of an event propagates among individuals is therefore essential for realistic evacuation simulations. Furthermore, individuals entering an environment for the first time often possess incomplete or inaccurate knowledge of exit locations. In emergencies, such individuals must rely on environmental cues, signage, authorities, or the behavior of nearby people to identify viable evacuation routes. Empirical studies indicate that, under high emotional stress, individuals may even forget the entrance through which they originally entered a building [2]. Consequently, modeling the acquisition and diffusion of exit knowledge within a crowd is a critical component of evacuation realism.

Emotional factors further shape evacuation dynamics. High levels of uncertainty and perceived personal risk generate fear, which directly influences decision-making, movement speed, and social behavior [3]. Emotional states, particularly fear, can rapidly spread among nearby individuals, even affecting agents who are initially unaware of the triggering event. Conversely, trusted authority figures or trained staff may exert a calming influence, reducing fear levels and stabilizing crowd behavior.

Beyond cognitive and emotional processes, individual differences in personality also play a critical role in shaping human responses under stress [4]. Empirical studies show that individuals differ significantly in their tendency to experience and sustain negative emotions, as well as in how they perceive and react to threatening situations. In particular, the Neuroticism trait – commonly associated with emotional instability has been linked to increased sensitivity to stress, anxiety, and fear, as well as prolonged negative emotional states [5]. These differences suggest that individuals exposed to identical conditions may exhibit substantially different emotional and behavioral responses. Despite this, personality is often either omitted or modeled in a simplified manner in evacuation simulations [6–8].

To capture these complex interactions, realistic evacuation simulations must integrate models of event awareness, exit knowledge, emotional dynamics, and personality, and explicitly link these internal states to agent decision-making and observable behavior. Our work builds upon and extends the ESCAPES evacuation framework [1], by introducing several cognitive, emotional, and personality-driven mechanisms that enable the modeling of heterogeneous human responses during emergencies.

The main contributions of this work are as follows:

- We propose a unified agent-based evacuation framework that integrates event awareness, exit knowledge, emotional dynamics, and personality into a coherent cognitive-behavioral model, enabling the interaction between perception, cognition, emotion, and behavior.
- We introduce a dynamic representation of exit knowledge as a memory process subject to acquisition, forgetting, and recall, enabling the modeling of confusion and disorientation under stress. The framework further incorporates parametric, agent-specific forgetting probabilities, allowing the representation of heterogeneous memory robustness linked to experience and training.

- We develop a continuous and graded mechanism for knowledge transfer between agents, where information about the event propagates through local interactions with varying influence depending on the certainty and source of information.
- We develop a continuous fear model in which panic emerges as a high-intensity state, driven by multiple sources including fire exposure, uncertainty, and stagnation. Fear directly affects perception, decision-making, movement, and physical interactions, allowing the emergence of realistic crowd phenomena such as falls, trampling, and injuries.
- We incorporate heterogeneous evacuation decision-making through individualized evacuation thresholds, capturing variability in agents' sensitivity to perceived risk.
- We integrate personality through the OCEAN model [9], focusing on the Neuroticism trait, modeling how stable individual differences systematically modulate fear generation, propagation, stagnation sensitivity and recovery dynamics, leading to heterogeneous emotional responses and evacuation outcomes even under identical conditions.

Finally, we evaluate the proposed framework through a series of controlled simulation experiments, demonstrating its impact on evacuation outcomes and its ability to reproduce key behavioral patterns observed in real emergency situations.

The rest of this paper is organized as follows. Section 2 reviews the related work on crowd evacuation modeling, with emphasis on the diverse factors shaping agent behavior – such as cognitive, emotional, social, physiological, and environmental influences – and on empirical studies examining early human responses during emergency events. Section 3 presents the proposed modeling framework, including the mechanisms for event awareness, exit knowledge, fear dynamics, and personality modulation. Section 4 describes the experimental methodology, simulation environment, and evaluation metrics. Section 5 presents and analyzes the results of the simulation experiments. Finally, Section 6 discusses the implications and limitations of the proposed framework, and Section 7 concludes outlining several directions for future research.

# 2 Related Work

## 2.1 Factors affecting Agent behavior

Agent-based simulations have emerged as the predominant methodological framework for crowd modeling, particularly when the objective is to capture heterogeneity, autonomy, and socially situated decision-making at the individual level. By modeling crowds as collections of interacting agents with internal states and goal-driven behaviors, this paradigm enables complex collective dynamics to emerge from local interactions. A comprehensive survey by Khan and Deng [4] systematically reviews the state of the art in agent-based crowd simulation, highlighting the wide range of factors that researchers have sought to integrate in order to enhance the behavioral authenticity of simulated crowds.

Modeling a heterogeneous crowd is inherently complex, as individual behavior in emergency situations is shaped by the interplay of psychological, emotional, physiological, social and environmental factors. Consequently, a substantial body of research has focused on incorporating these dimensions into agent-based models to move beyond over simplistic representations of evacuees.

Locomotion policies, while fundamental for movement realism, are only one component of agent behavior. Traditional approaches to crowd navigation include social force models, rule-based systems, and cellular automata, while more recent work explores reinforcement learning [10], data-driven methods, and hybrid approaches incorporating game theory, fuzzy logic, or vision-based perception. As an extensive discussion of these approaches is beyond the scope of this work, interested readers are referred to Khan and Deng [4] for a comprehensive review of both classical and modern locomotion strategies in crowd simulation.

One major line of work concerns personality modeling, where agents are endowed with stable individual traits that influence perception, decision-making and interaction. Widely adopted frameworks such as the OCEAN [9] and PEN [11] personality models have been used to introduce systematic behavioral variability among agents. Differences in traits such as extraversion, neuroticism, or conscientiousness have been shown to affect risk perception, social engagement, leadership tendencies, and responses to crowding, thereby producing more natural and diverse behavioral patterns in simulated crowds.

Beyond static personality traits, researchers have emphasized the importance of emotions in shaping emergent behavior, particularly in emergency contexts. Emotion models grounded in cognitive appraisal theories and social psychology, such as the OCC [12] framework, have been integrated to allow agents' emotional states to evolve dynamically in response to perceived events [13,14]. These models typically link emotions to action selection, enabling behaviors such as panic, hesitation, helping, or aggression to arise from internal states rather than being hard-coded. Several approaches further bridge personality and emotion through intermediate representations, such as PAD-based mappings [15], enabling consistent transitions between traits, emotional responses, and observable actions.

Closely related to emotion modeling is research on stress, perception, and emotion contagion. Appraisal-based approaches [16,17] assume that agents continuously evaluate environmental stimuli relative to their goals and well-being, leading to subjective emotional responses. Stress models have been employed to simulate how increasing threat or discomfort alters movement speed, attention, and decision-making [17,18]. To capture social influence, various emotion contagion mechanisms have been proposed, ranging from epidemiological models inspired by disease spread [19–21] to thermodynamics-based formulations [22] and group-level emotion representations. Tsai et al. [1] introduced a multi-agent evacuation simulation framework, ESCAPES, in which agents are capable of adopting the emotional state of others, particularly those exhibiting the strongest effective intensity or holding a distinctive social role. Minh et al. [23] proposed an emotion model that captures both the temporal evolution and the transmission of emotions within a crowd, embedding it into an evacuation framework. Their approach accounts for when emotions are triggered, how their intensity changes over time, and how they spread between individuals, including the mechanisms through which emotions are expressed, perceived, and subsequently influence the recipient's behavior. Similarly the ASCRIBE model [24] conceptualizes emotion contagion as a continuous, group-level process within a multi-agent system, treating emotion as a collective phenomenon rather than solely an individual attribute. Cho et al. [25] incorporated the Belief-Desire-Intention (BDI) architecture into crowd simulation. Under this paradigm, agents act based on internal desires (goals), informed by beliefs (their understanding of the environment), and execute intentions (planned actions) aimed at achieving those goals, thereby enabling more adaptive and cognitively grounded behavior in dynamic settings. These approaches aim to model how panic or calmness propagates through a crowd and how collective emotional states emerge from local interactions.

Another critical set of factors relates to the physical environment and spatial constraints. Geometric features such as exit width, bottlenecks, visibility, stairways, and obstructions have been shown to strongly influence evacuation efficiency and crowd safety. Numerous studies [26,27] explicitly model these constraints to investigate how spatial design interacts with individual behavior, congestion, and panic.

In addition to individual cognition and environment, social structure and group dynamics have received increasing attention. Models incorporating leader-follower relationships [28–30], social comparison processes [31], and social distancing [32,33] capture how individuals adjust their behavior based on nearby others, particularly under uncertainty. Other approaches explicitly model family groups, trained leaders, or social roles to reflect realistic evacuation scenarios [34].

Researchers have also explored physiological factors to account for physical limitations and variability in movement. Attributes such as age, health, fatigue, body shape, and physical strength have been incorporated to influence walking speed, collision handling, and endurance under high-density conditions [35–37]. These physiological considerations enhance realism by acknowledging that not all agents are equally capable of sustaining rapid or forceful movement during evacuations.

Finally, several models emphasize roles, goals, and needs as central drivers of agent behavior [38–42]. Rather than treating agents as homogeneous evacuees, these approaches assign individuals distinct roles, daily routines, and personal objectives, enabling more nuanced action selection. Role-based and need-driven models allow agents to switch behaviors in response to context, social cues, and internal states, producing richer and more believable crowd dynamics.

Despite these advances, integrating all these dimensions into a unified and computationally tractable framework remains a significant challenge. Many existing simulations focus on a subset of factors, leaving open questions regarding how best to combine psychological, emotional, social, physiological, and environmental components within a single coherent framework, as well as how to systematically evaluate the realism of the resulting behaviors [4].

## 2.2 Early Behavioral Responses to Evacuation Emergencies

There is an increasing demand for computational models capable of simulating crowd dynamics and evacuation processes in emergency situations [43]. Such models are widely used to inform evacuation planning, public safety policies, and the design of transport infrastructures. However, their effectiveness critically depends on how accurately they capture real human behavior under threat [44]. A persistent gap has been identified between evacuation simulations and actual evacuee behavior, which substantially limits the predictive and practical value of these models.

Many evacuation models rely on simplifying assumptions that poorly reflect empirical observations. Agents are often assigned uniform walking speeds and deterministic exit choices [45], are assumed to immediately recognize the emergency and head directly towards an exit, and are modeled as socially independent decision-makers. Additionally, several simulation frameworks presume predominantly competitive behavior, frequently leading to the well-known “faster-is-slower” effect. While these assumptions facilitate model implementation, they fail to represent the diversity, social embededness, and contextual sensitivity of human behavior during real emergencies.

As a result, understanding what evacuees actually do in the critical moments immediately following emergency onset is essential for developing realistic evacuation models and improving emergency response strategies.

A substantial body of literature has documented a wide range of behavioral responses during emergencies. Early physiological theories describe the “fight-or-flight” response as a basic reaction to perceived danger, whereby individuals either confront the threat or attempt to escape from it [45]. However, subsequent research has demonstrated that human behavior in emergencies is far more heterogeneous.

One commonly observed response is freezing, where individuals experience cognitive paralysis under extreme stress and temporarily fail to act [46]. Others exhibit continuation of routine activities, such as retrieving belongings, remaining seated, or following habitual routes despite immediate danger [47–49]. Affiliative behavior, in which individuals move towards familiar people or locations rather than directly away from danger, has also been widely reported [50].

Social behaviors play a central role in emergency contexts. Numerous studies show that evacuees often seek cues from others, adjusting their actions based on nearby individuals [51,52]. Information seeking and sharing are common, with evacuees frequently exchanging guidance with both familiar and unfamiliar others [53]. Importantly, information is more likely to be accepted when it originates from a credible source or a perceived authority [54,55].

Both pro-social and anti-social behaviors have been documented. While competitive actions such as pushing or prioritizing oneself can occur, a growing body of evidence highlights widespread cooperation, orderly queuing, and helping behaviors, even under severe threat [56–58]. Experimental and observational studies further suggest that evacuees often leave via familiar exits, but this tendency can be overridden when alternative exits are open and the outside can be seen [59], visibly safer or closer [60].

Finally, crowd density and environmental constraints strongly shape evacuation dynamics. Even at extreme densities, individual movement rarely ceases entirely; instead, congestion can give rise to phenomena such as “crowd turbulence”, increasing the risk of displacement and trampling [61,62]. Empirical observations also contradict the assumption of uniform movement, showing acceleration patterns and adaptive behaviors depending on proximity to exits and environmental visibility [63].

Despite extensive research, much of what is known about evacuee behavior is derived from retrospective self-reports, survivor interviews, or controlled experimental studies. While valuable, these sources suffer from notable limitations. Self-reports are prone to memory distortions and post-hoc rationalization, whereas experiments often lack ecological validity, as ethical constraints prevent researchers from exposing participants to genuine life-threatening danger. Consequently, there remains considerable uncertainty regarding how individuals actually behave in the immediate aftermath of real emergencies.

Philpot and Levine [43] address this gap by conducting what they believe to be the first video-based analysis of evacuee behavior following a real-life explosion in a subway train carriage. Using surveillance camera footage from an actual public transport emergency, their study provides rare, high-fidelity observations of behavior in the seconds immediately following emergency onset. Their findings reveal substantial variation in initial responses. Some individuals ran or walked away from the explosion, while others froze, observed the scene for extended periods, retrieved belongings, hid, or deliberately remained seated to allow others to pass. Both pro-social and anti-social behaviors were observed; however, pro-social actions – such as yielding space, waiting patiently, and protecting vulnerable individuals – were more prevalent than competitive ones. Notably, running behavior exhibited strong spatial clustering, with individuals more likely to run if their immediate neighbors were also running. Distance from the explosion itself was not statistically associated with running, highlighting the importance of social influence over purely physical factors. Exit choice further reflected this pattern: although prior literature suggests evacuees typically use the closest exit, most individuals in this incident evacuated via the exit furthest from the explosion, regardless of familiarity or proximity. This behavior is consistent with both a desire to maximize distance from the threat and a “follow-the-majority” effect driven by early movers. Overall, the study provides compelling evidence that evacuee behavior is highly social, context-dependent, and often cooperative.

# 3 Framework Description

## 3.1 Modeling Awareness and Certainty of an Emergency Event

Event knowledge is a fundamental prerequisite for evacuation decision-making. Unlike simplified evacuation models, where agents react immediately to the presence of a hazard, agents in our framework initiate evacuation only after directly perceiving the emergency or accumulating sufficient information to be convinced that the event is real and requires action. This approach is consistent with evidence accumulation theories on decision-making under uncertainty, according to which individuals gather information until a subjective confidence threshold is reached before committing to an action [64]. It also captures the well-documented pre-movement phase in evacuations, during which individuals often delay response despite perceiving alarms or cues, leading to increased overall evacuation time [65,66].

Our approach builds upon the ESCAPES framework, which explicitly models the gradual acquisition and social transmission of event knowledge. We extend this framework to incorporate richer representations of uncertainty, individual heterogeneity in decision-making, and graded social influence.

In contrast to ESCAPES, where event certainty is represented using a small set of discrete levels, we adopt a continuous representation of certainty. This enables smoother belief updates and more realistic transitions from uncertainty to conviction.

Each agent maintains a continuous internal state variable, termed ***Event Certainty Level (ECL)***, defined in the range [0, 1]. This variable represents the agent's subjective belief regarding the occurrence of an emergency event (e.g., a fire). A value of 0 corresponds to complete unawareness, intermediate values indicate partial or uncertain knowledge, and a value of 1 reflects absolute certainty, typically resulting from direct perception of the event.

An agent initiates evacuation when its ECL exceeds an individual-specific threshold. Formally, the evacuation decision of agent $i$ is defined as:

$$Evacuate_i = \begin{cases} 1, if\ ECL_i \geq \theta_\iota \\ 0, otherwise \end{cases}$$

where $ECL_i$ denotes the Event Certainty Level of agent $i$, and $\theta_\iota \in [0,1]$ is the evacuation threshold. Lower threshold values of $\theta_i$ correspond to individuals more willing to initiate evacuation under uncertainty, whereas higher values indicate a need for stronger confirmation before committing to evacuation.

While the threshold is defined over a continuous domain to support fine-grained heterogeneity, representative values can be used to capture typical behavioral tendencies. For example, low thresholds (e.g., $\Theta \approx 0.2$) correspond to highly responsive individuals who react quickly to emerging threats, medium thresholds (e.g., $\Theta \approx 0.5$) represent average responsiveness, and high thresholds (e.g., $\Theta \approx 0.8$) correspond to more conservative individuals who delay action until strong confirmation is obtained. These indicative values are used in the experimental analysis to model heterogeneous populations, while preserving the generality of the formulation. The threshold therefore captures individual responsiveness to perceived risk. Empirical evidence suggests that such differences may be influenced by prior experience with hazardous events [54] or risk denial [43], leading to faster or more cautious responses when facing threats under uncertainty. Assigning heterogeneous thresholds across agents allows diverse evacuation behaviors to emerge naturally. All agents are initialized with zero ECL, reflecting the absence of prior knowledge:

$$ECL_i(0) = 0, \forall i$$

The evolution of event certainty is modeled as an event-driven update process, triggered by two types of events:

1. Direct environmental perception
2. Local interaction with neighboring agents

The value of ECL changes only when such events occur.

Direct perception provides the strongest evidence: agents continuously monitor their surroundings within a predefined vision range, and unobstructed observation of the fire immediately sets ECL to its maximum value, enabling instant evacuation. Direct perception induces an instantaneous update:

$$If\ S_i = 1 \Longrightarrow ECL_i \longleftarrow 1$$

where $S_i \in \{0,1\}$ denotes whether agent $i$ has directly seen the fire.

In addition to direct perception, agents acquire information through social interaction. Agent $i$ updates its ECL through pairwise interaction events occurring when another agent $j$ enters its interaction range. At each such event, the ECL is updated as:

$$ECL_i \leftarrow \min\left(1, max\left(0, ECL_i + \delta_j\right)\right)$$

where $\delta_j$ denotes the influence of neighbor $j$ and depends on the informational state of neighbor $j$:

$$\delta_j = \begin{cases} a_s, & if\ S_j = 1 \wedge ECL_j \geq \theta_j \\ a_m, & if\ S_j = 0 \wedge H_j = 1 \wedge ECL_j \geq \theta_j \\ a_w, & if\ S_j = 0 \wedge H_j = 1 \wedge ECL_j < \theta_j \\ a_n, & if\ S_j = 0 \wedge H_j = 0 \end{cases}$$

where:

- $H_j \in \{0,1\}$ denotes whether agent $j$ has heard about the fire
- $\theta_j$: evacuation threshold of agent $j$
- $a_s > a_m > a_w > 0 > a_n$ : influence coefficients

In the implemented model, these coefficients are instantiated as:

$$a_s = 0.2, a_m = 0.1, a_w = 0.05, a_n = -0.001$$

Unlike the categorical interaction rules used in ESCAPES, this formulation captures graded social influence, where both the certainty level and the source of information determine the strength of influence:

- Agents who have directly perceived the fire and are convinced exert strong influence
- Agents who are convinced but rely on indirect information exert moderate influence
- Agents who are uncertain exert weak influence
- Agents with no knowledge introduce slight negative influence, modeling skepticism or reassurance effects.

Unlike the categorical interaction rules used in ESCAPES, the proposed formulation introduces graded social influence, where the impact of an interaction depends on both the neighbor's level of certainty and the source of their information. Information from agents who have directly perceived the event produces a strong increase in certainty, whereas interactions with uncertain or uninformed agents result in weaker or even negative updates.

This formulation is supported by empirical findings showing that evacuation decisions are strongly influenced by nearby individuals. People are significantly more likely to evacuate when they observe others doing so, while the presence of passive individuals may delay evacuation response [44].

## 3.2 Modeling Exit Knowledge

Incomplete and uncertain knowledge of exit locations is a critical cognitive factor in evacuation, directly affecting both pre-evacuation delay and overall efficiency. Empirical evidence suggests that under stress, individuals may fail to recall known exits, select suboptimal routes, or rely on social cues instead of spatial memory [43]. Motivated by these observations, we model exit knowledge as a dynamic, agent-specific cognitive state that evolves over time and interacts with perception, emotional state, and social behavior. This extends the ESCAPES framework, where exit knowledge is represented primarily as a binary attribute indicating whether an agent knows a specific exit. In the proposed model, each agent $i$ maintains a set of exit knowledge entries

$$E_i = \{e_{i,1}, e_{i,2}, \cdots e_{i,n}\}$$

where each entry corresponds to a specific exit $x$. Exit knowledge is represented as a structured tuple

$$e = (x, k, f, w, t)$$

where $x$ denotes the exit, $k \in \{0,1\}$ indicates whether the exit has been learned, $f \in \{0,1\}$ denotes temporary forgetting, $w > 0$ is an importance weight reflecting cognitive salience of the exit, and $t$ records the last time the exit was visually perceived. This representation separates knowledge acquisition from recall. An agent may therefore know an exit ($k = 1$) but be temporarily unable to recall it ($f = 1$), capturing stress-induced cognitive degradation rather than permanent information loss.

### 3.2.1 Initial exit knowledge

At initialization, all agents know a primary exit corresponding to their entry location. This exit is assigned a lower importance weight, reflecting its lower susceptibility to forgetting due to higher familiarity. Empirical studies show that individuals tend to evacuate via familiar routes – especially the path of entry- rather than distributing across available exits. This behavior has been linked to habitual navigation and learned irrelevance, whereby frequently observed but rarely used emergency exits may be cognitively discounted [67,68]. Assigning higher cognitive salience to the entry exit captures this effect.

Knowledge of secondary exits is modeled probabilistically based both on environmental familiarity and prior exposure to evacuation-related information or training. Agents are categorized into familiarity classes

$$A_i \in \{Staff, RegularVisitor, FirstTimeVisitor\}$$

which represent different levels of spatial familiarity. Empirical evidence suggests that familiar individuals (e.g., staff) respond more proactively by quickly identifying exits or guiding others, whereas first-time visitors are more likely to exhibit reactive behavior and wait for instructions or external guidance [69].

Instead of fixed probabilities, we introduce a configurable parameter $P_{init}(A_i)$, representing the probability that an agent of type $A_i$ initially knows a given secondary exit. This allows the framework to capture variability due to training, institutional policies, or information availability, unlike ESCAPES where such probabilities are implicit.

For example, in well-trained environments, staff may have $P_{init}(A_i) \approx 1.0$, whereas in less structured settings lower values (e.g., 0.8-0.9) may be more appropriate. Similarly, first-time visitors may possess partial knowledge when information is clearly communicated (e.g., signage or maps). This parameterization enables the framework to represent the effects of organizational preparedness and communication on evacuation behavior [69].

### 3.2.2 Dynamic acquisition of exit knowledge

During the simulation, agents may acquire or reinforce exit knowledge through direct visual perception. While ESCAPES determines exit knowledge primarily at initialization, our framework allows continuous updates. When an exit enters an agent's visual range and is confirmed via raycasting (simple observation), the corresponding entry is updated as

$$k_{ix} \leftarrow 1, \qquad f_{ix} \leftarrow 0, \qquad t_{ix} \leftarrow t_{now}$$

Thus, perception both establishes exit knowledge and restores recall, allowing agents to transition dynamically between awareness and temporary disorientation.

### 3.2.3 Forgetting and recall

Exit recall is modeled as a probabilistic process influenced by cognitive, emotional, and temporal factors, including agent type, fear level, time since last visual contact, and exit importance, unlike ESCAPES, where forgetting is implemented as a single stochastic event affecting knowledge of the entry exit. Higher fear and longer periods without visual reinforcement increase forgetting likelihood, consistent with trace decay and retrieval failure mechanisms [70].

For a given exit entry e, the probability of temporary forgetting is

$$P_{forget}\,(e) = \min\big(1, \max(0, B(A_i) \cdot (1 + F_i) \cdot M_t(e) \cdot w)\big),$$

where $B(A_i)$ is a base forgetting probability determined by agent type, $F_i \in [0,1]$ denotes the agent's fear level, and $M_t(e)$ is a time-dependent multiplier defined as:

$$M_t(e) = 0.5 + 1.5 \cdot min\left(1, \frac{t_{now} - t}{T_{decay}}\right)$$

The multiplier $M_t(e)$ increases linearly from 0.5 to 2.0 as the time since last visual confirmation approaches a predefined decay horizon $T_{decay}$ , after which it remains constant. The importance weight $w$ modulates susceptibility to forgetting, with lower values corresponding to exits that are more cognitively salient.

Unlike the ESCAPES model, where exit forgetting is represented as a simple stochastic event, our formulation is grounded in empirical findings showing that stress can impair recall and delay activation of learned responses. Under high stress, individuals may experience temporary cognitive impairment that disrupts recall or temporary "freeze" states, particularly in the absence of prior training [46].

Consequently, agents with greater experience or training are assigned lower base forgetting probabilities, reflecting more stable cognitive schemas. For example, staff agents may have near-zero forgetting probabilities, representing reliable recall under stress.

### 3.2.4 Exit selection during evacuation

Upon deciding to evacuate, an agent selects an exit from the set of currently recalled exits:

$$E_i^{active} = \{e \in E_i | k = 1 \wedge f = 0\}$$

If this set is non-empty, the agent moves toward the nearest exit. Otherwise, it enters a temporary disorientation state and relies on social or exploratory strategies.

Unlike ESCAPES, where exit knowledge is transmitted explicitly, agents in our framework may exploit others' knowledge indirectly through leader-following behavior. Specifically, an agent first attempts to follow a nearby agent who is evacuating and demonstrably recalls an exit, reflecting reliance on knowledgeable individuals in unfamiliar environments [69].

If no such agent is available, the agent engages in stochastic exploration to discover exits or encounter informed neighbors. This behavior is consistent with findings that uncertainty promotes information seeking actions, such as wandering or environmental scanning, to reduce ambiguity and identify viable routes [64].

## 3.3 Fear Modeling

Fear represents the emotional state of an agent during evacuation and acts as a key mediator between perception, cognition, and behavior. Panic is modeled as an intensified stress response to perceived danger, often triggered by uncertainty and evolving threat conditions [71,72]. This conceptualization aligns with the notion of psychological entropy, where uncertainty induces anxiety and drives adaptive behavioral responses aimed at reducing ambiguity [64]. Consistent with prior evacuation models, perceived fire risk dynamically influences agent's psychological state and behavior [73].

Within this framework, fear directly affects motion stability, decision-making, and susceptibility to physical interactions, enabling the emergence of panic-driven crowd phenomena. Previous work shows that increased stress influences movement dynamics, including speed and escape behavior [73,74].

The concept of panic in crowd dynamics remains theoretically ambiguous, encompassing a wide range of phenomena, such as extreme fear, erratic motion, and non-cooperative or competitive interactions [44]. This lack of definitional consensus poses a fundamental challenge for evacuation modeling, as any representation of panic necessarily relies on model-specific assumptions.

To address this ambiguity, we model panic as a high-intensity state within a continuous fear spectrum rather than as a discrete category. Each agent $i$ maintains a continuous Fear Level $F_i(t) \in [0,1]$, updated at every simulation step. Unlike ESCAPES, where fear is discrete ($FearFactor \in \{0,1,2\}$) primarily affecting movement speed, this formulation enables gradual emotional escalation,  and agent heterogeneity, consistent with interpretations of panic as an extreme manifestation of fear rather than a qualitatively distinct behavioral state [44]. This is further supported by studies highlighting the difficulty of discretely categorizing transitions from calm to panic and supporting continuous or multi-level representations of emotional states [71]. Similarly, Cao et al. [73] model stress as a continuous variable evolving with environmental conditions, showing that evacuees transition gradually between low and high stress states depending on their proximity to fire hazards.

Extreme panic behaviors are relatively rare in real evacuations [75]. Accordingly, panic in our model emerges only when Fear Level exceeds a high threshold, capturing infrequent but critical breakdowns in rational behavior. This distinction aligns with findings showing that moderate panic may facilitate evacuation, whereas excessive panic impairs cognition and leads to maladaptive behavior [71].

Rather than treating fear as a single abstract variable, we model it as the result of multiple concurrent stressors related to environmental danger, informational uncertainty, motion effectiveness, and social influence. While ESCAPES considers several situational factors influencing fear – such as proximity to the event, authority presence, and neighbors' fear levels – these influences are implicitly merged into a single Fear Factor variable. In contrast, the proposed framework explicitly decomposes fear into multiple interpretable components. At each update cycle, the agent evaluates distinct fear components and combines them using a dominance-based rule, ensuring that the strongest perceived threat governs emotional response. This dominance-based aggregation is consistent with approaches reported in the emotion contagion literature, where emotional intensity is determined by the most influential contributing factor rather than by additive accumulation [76]. Formally, fear is computed as

$$F_i(t) = max\left(F_i(t-1), F_i^{fire}(t), F_i^{internal}(t)\right)$$

where $F_i^{fire}$ denotes fire-induced fear, and $F_i^{internal}$ aggregates cognitive and motion-related stressors.

Fire-induced fear models both direct and indirect exposure to danger. Agents without information about the fire experience no fear. Once information is acquired, fear intensity depends on perceptual source, distance to the fire, and certainty regarding the need to evacuate. Agents who directly perceive the fire experience elevated fear, modulated by spatial proximity: unsafe distances produce high fear, whereas safe-distance observations yield moderate fear. Agents who only receive indirect information exhibit fear driven primarily by uncertainty: unconvinced agents experience low fear, while convinced agents exhibit higher fear due to the lack of spatial grounding of the threat. This formulation aligns with stress-based evacuation models in which psychological response is explicitly driven by proximity to the fire and perceived threat intensity [73], where stress increases as evacuees approach the fire and decreases when moving to safer distances.

Fire-induced fear is treated as a high-priority stimulus. When sufficiently intense, it overrides any calming influence exerted by authority figures, modeling the dominance of immediate perceived danger over social reassurance. In ESCAPES, proximity to the event is also one of the factors influencing the *FearFactor*. However, the model does not explicitly distinguish between perceptual sources of threat. Our formulation therefore separates direct perception from socially transmitted information, allowing fear intensity to depend on both distance and perceptual certainty.

Uncertainty fear captures anxiety arising from lack of actionable information and is calculated only after an agent has committed to evacuation. If an agent recalls at least one exit, uncertainty fear is zero. If no exit is recalled but the agent follows a trusted leader, uncertainty is reduced but remains non-zero, reflecting reliance on external guidance. When neither exit knowledge nor leadership is available, uncertainty reaches high levels, producing strong emotional stress driven by disorientation and loss of control.

Stagnation fear models emotional escalation caused by ineffective evacuation progress. For agents that know an exit, fear increases when measurable progress toward that exit is not achieved. Progress is assessed objectively by monitoring reductions in distance to the exit and verifying actual movement. If the agent remains stationary or fails to reduce distance over time, stagnation fear grows gradually. For agents without any known exit, stagnation fear reduces to a time-based uncertainty process: fear increases monotonically as time passes without directional progress. Both cases reflect frustration, helplessness and perceived entrapment, which are well-documented contributors to panic escalation.

Both uncertainty and stagnation fear are supported by empirical findings from spatial cognition and way finding research indicating that the lack of environmental feedback – such as the inability to confirm progress or the absence of expected cues – generates uncertainty, increasing cognitive load and anxiety [64]. This aligns with the concept of psychological entropy, where uncertainty is inherently stress-inducing and motivates adaptive responses. Consequently, insufficient directional feedback during evacuation can elevate stress levels and contribute to disorientation and loss of control.

Uncertainty-induced and stagnation-induced fear are combined into an internal fear component using a dominance rule

$$F_i^{internal}(t) = max\left(F_i^{uncertainty}(t), F_i^{stagnation}(t)\right)$$

Fear spreads socially through local interactions. When agents encounter neighbors, they adopt the maximum fear level observed among them. This mechanism follows the baseline emotional contagion rule originally implemented in the ESCAPES simulator, where agents inherit the highest fear level among nearby individuals.

Rather than explicitly modeling herding as a predefined “follow-the-majority” rule, our approach captures social influence implicitly through local emotional contagion. This choice is consistent with findings in the literature indicating that herding is not a universal behavior. but a context-dependent response that emerges primarily under conditions of high uncertainty and limited environmental information [44]. This perspective is further supported by studies showing that panic contagion and social influence intensify under uncertainty and environmental ambiguity [71]. Moreover, empirical evidence suggests that individuals are more strongly influenced by nearby agents than by the crowd as a whole, and that imitative behavior becomes more pronounced when reliable personal knowledge or clear environmental cues are lacking [44].

By relying on local interactions, the framework allows imitative patterns to emerge naturally when agents are uncertain or disoriented, without enforcing global crowd-following behavior. This enables the coexistence of independent decision-making in informed agents and collective dynamics in uncertain groups, providing a more flexible and realistic representation of social influence in evacuation scenarios.

To prevent agents from remaining indefinitely in a panic state, a panic cooldown mechanism is introduced. If an agent remains panicked beyond a predefined duration, it enters a temporary cooldown phase. During this phase, fear is immediately reduced, panic cannot be re-triggered, and fear transmission from neighbors is disabled. After the cooldown period ends, the agent re-enters the normal decision-making flow. This mechanism introduces explicit temporal dynamics in the evolution of fear. Conceptually, this temporary recovery phase is analogous to the “recovered” state used in SIRS epidemiological models of emotional contagion, where individuals may become temporarily resistant to further emotional influence before returning to a susceptible state [71,77]. In ESCAPES, fear levels may change abruptly due to interactions with authority figures without explicitly modeling gradual emotional recovery.

## 3.4 Effects of Fear and Panic on Agent Behavior

While the previous subsection focused on the internal dynamics of fear generation and propagation, fear in our framework is not merely an abstract emotional variable. Instead, it directly shapes agents’ movement, decision-making, and physical interactions during evacuation. This subsection describes how fear and panic states translate into concrete behavioral effects.

In our framework, fear is treated as a driving force that directly alters perception, cognition, motion, and physical interaction [72]. This contrasts with the ESCAPES formulation, where fear primarily modifies movement speed and influences probabilistic behavior selection. Elevated fear levels progressively degrade the agent’s ability to perceive the environment, recall knowledge, make rational decisions, and maintain controlled movement. When fear exceeds a critical threshold, agents enter a panic state that fundamentally alters their behavioral and physical dynamics, leading to emergent crowd phenomena such as collisions, falls, trampling, injuries, and fatalities. This modeling choice is supported by findings in the literature suggesting that panic can act as a major contributor to injuries during emergency situations and may significantly aggravate the harmful consequences of the underlying hazard [44,71].

### 3.4.1 Effect on Movement Speed and Control

Fear influences both the speed and stability of agent movement. In the ESCAPES framework, higher fear levels primarily increase movement speed in order to accelerate escape behavior. In contrast, our framework associates high fear levels with reduced coordination and behavioral instability. As fear increases, agents attempt to move faster in an effort to escape perceived danger. However, once fear surpasses a panic threshold, increased urgency is accompanied by a loss of motor control.

In the panic state, instead of controlled navigation, the agents exhibit:

- Unstable or erratic movement
- Delayed responsiveness to direction changes
- Reduced ability to avoid obstacles and nearby agents

This behavioral shift is crucial for realism, as panic does not simply increase speed – it decreases coordination and decision quality.

This modeling assumption is consistent with literature linking panic to competitive and non-cooperative behavior, where individuals push, compete for space, and deviate from socially coordinated norms [44]. Such dynamics lead to emergent crowd phenomena – including bottlenecks, falls, and trampling – that can significantly impact evacuation outcomes and increase the risk of injury, particularly under high-stress conditions [69,71].

### 3.4.2 Effect on Perceptual Capability

Fear impairs perceptual processing, and in panic states agents may fail to detect exits within their field of view. This effect is consistent with evidence that heightened arousal narrows attentional focus and reduces processing of peripheral environmental cues, leading to attentional tunneling in high-stress situations [64,72,78]. As a result, agents may overlook visible exits, increasing disorientation and reliance on non-cognitive behaviors such as blind escape or crowd following.

### 3.4.3 Effect on Memory and Exit Recall

In our framework, fear impairs recall of previously learned exit knowledge, modeling stress-induced memory degradation rather than permanent information loss. The probability that an agent temporarily forgets a known exit increases monotonically with its fear level. Importantly, forgetting is modeled as reversible: an agent may re-establish recall only when fear subsides below the panic threshold. This modeling approach is consistent with findings showing that elevated arousal disrupts memory encoding and retrieval [64].

### 3.4.4 Effect on Rational Decision-Making

Fear plays a central role in determining whether an agent is capable of goal-directed and rational decision making. Unlike ESCAPES, where fear is tightly coupled with probabilistic behavior selection within a BDI-style framework, our framework treats fear as a global emotional regulator that influences evacuation indirectly through its effects on cognition, perception, and movement stability. When an agent has directly perceived the fire and is within an unsafe distance, it adopts an avoidance behavior that depends on its fear level:

- $FearLevel < 0.7$: the agent behaves rationally, combining fire avoidance with route planning toward a known exit.
- $FearLevel > 0.7$: rational planning is suppressed. The agent no longer evaluates exits or navigation alternatives and instead moves directly away from the fire, exhibiting reactive escape behavior.

This transition from goal-directed navigation to instinctive flight reflects the behavioral shift observed under high stress, where individuals prioritize immediate threat avoidance over planned actions, consistent with interpretations of panic as a non-deliberative state [44,73]. At the same time, the model distinguishes panic from suboptimal but still rational behavior under uncertainty, as actions such as fleeing or rapid movement may remain reasonable given limited information [79,80]. Accordingly, agents in our framework retain goal-directed behavior below the panic threshold and switch to non-deliberative responses only when this threshold is exceeded.

#### 3.4.5 Physical Consequences of Panic: Collisions, Falls, and Tramping

Panic fundamentally alters how agents interact physically with their environment and with each other. The loss of obstacle avoidance and controlled motion increases the frequency and severity of collisions. Strong collisions may cause agents to fall, immediately immobilizing them and triggering injury evaluation.

Trampling damage accumulates continuously and is amplified by the presence of nearby panicking agents, reflecting the increased force and instability generated by panic-driven crowds. Sustained trampling can result in severe injury or death, permanently removing agents from the simulation.

### 3.5 Personality Modeling: Neuroticism-Driven Modulation of Fear Dynamics

To incorporate individual differences in emotional response, we adopt a personality representation based on the OCEAN model [9]. The personality of each agent $i$ is defined as a vector:

$$\varphi_i = [\psi_i^O, \psi_i^C, \psi_i^E, \psi_i^A, \psi_i^N]$$

where $\psi^O$ denotes openness, $\psi^C$ conscientiousness, $\psi^E$ extraversion, $\psi^A$ agreeableness, and $\psi^N$ neuroticism. Each trait is modeled as a continuous variable in the range $[0,1]$, allowing heterogeneous personality profiles across agents.

According to the OCEAN model, Neuroticism (also referred to as emotional instability) reflects the tendency of an individual to experience negative emotions such as anxiety, fear, frustration, and stress, as well as the degree to which such emotional responses persist over time [5]. Individuals with high Neuroticism are more likely to perceive situations as threatening and exhibit stronger and longer-lasting emotional reactions, whereas individuals with low Neuroticism tend to remain calm and emotionally stable under stress. In the present work, we focus exclusively on the role of Neuroticism ($\psi_i^N$) and its impact on the emotional dynamics of agents. The remaining personality traits are included in the representation but are not yet integrated into the framework and are left for future work.

Rather than introducing new behavioral rules or explicitly mapping personality traits to predefined behavior patterns, as in prior work where personality directly determines behavioral modes (e.g., leader, follower, or irrational agent) or modifies motion and decision rules [6,7], Neuroticism in our framework is implemented as a modulation factor that systematically alters multiple stages of the fear generation and propagation process. This approach allows personality to influence evacuation dynamics indirectly, through its effect on emotional sensitivity, escalation, and recovery. Specifically, Neuroticism is modeled as a multi-level modulation mechanism that affects:

- The amplification of fire-induced and uncertainty-driven fear,
- The rate of stagnation-induced emotional escalation,
- Susceptibility to emotional contagion, and
- the duration and effectiveness of panic recovery.

#### 3.5.1 Modulation of Fire-Induced and Uncertainty Fear

As described in section 3.3, fear is computed through distinct components, including fire-induced fear $F_i^{fire}(t)$ and uncertainty-induced fear $F_i^{uncertainty}(t)$. These components are first evaluated independently, producing each a baseline fear value:

$$F_i^{base}(t) \in [0,1]$$

To incorporate personality effects, we introduce a panic activation threshold $\tau_p \in (0,1)$. Neuroticism modulates fear only when baseline fear remains below this threshold, amplifying emotional response without exceeding the panic boundary. The modulated fear is defined as:

$$F_i(t) = \begin{cases} F_i^{base}(t) + \psi_i^N \cdot \left(\tau_p - F_i^{base}(t)\right), if\ F_i^{base}(t) < \tau_p \\ F_i^{base}(t),\ \ otherwise \end{cases}$$

This formulation increases sensitivity to perceived threat in a controlled manner. Agents with higher Neuroticism experience stronger emotional reactions under identical environmental and informational conditions, while preserving the interpretation of panic as a threshold-driven emergent state.

### 3.5.2 Modulation of Stagnation-Induced Fear

In addition to fire and uncertainty components, Neuroticism also affects stagnation-induced fear $F_i^{stagnation}(t)$, which captures emotional escalation due to lack of progress during evacuation.

In the baseline formulation, stagnation fear increases gradually when agents fail to reduce their distance to a target exit or remain stationary. To incorporate personality effects, Neuroticism modulates the rate of this temporal escalation.

Specifically, higher values of $\psi_i^N$ accelerate the growth of stagnation-induced fear, causing agents to interpret lack of progress as more threatening. As a result, high-Neuroticism agents reach elevated fear levels and panic states more rapidly under conditions of congestion or blocked movement.

This mechanism captures the increased sensitivity of certain individuals to perceived inefficiency or entrapment, leading to faster emotional deterioration in crowded or uncertain environments.

### 3.5.3 Modulation of Emotional Contagion

Fear propagation through social interaction is also influenced by Neuroticism. In the baseline formulation, agents adopt the maximum fear level observed among nearby neighbors. In the extended model, this process becomes gradual and personality-dependent. When agent i interacts with a neighbor j with higher fear level, the update rule becomes:

$$F_i(t) \leftarrow F_i(t) + \psi_i^N \cdot \left(F_j(t) - F_i(t)\right)$$

This formulation causes the agent's fear to move toward the neighbor's fear level at a rate proportional to its Neuroticism. As a result, low-Neuroticism agents exhibit resistance to emotional contagion and maintain more stable internal states, whereas high-Neuroticism agents rapidly converge toward the fear levels of more distressed neighbors.

This modification replaces abrupt fear adoption with a continuous convergence process, enabling more realistic modeling of heterogeneous susceptibility to social influence.

### 3.5.4 Modulation of Panic Duration and Cooldown Dynamics

Neuroticism further affects the temporal dynamics of panic and recovery through modification to the panic cooldown mechanism.

**Time before entering cooldown**. The duration an agent remains in the panic state before entering cooldown is extended proportionally to Neuroticism. Formally, the effective panic duration is:

$$T_{panic}^{eff} = T_{panic} \cdot \left(1 + \psi_i^N\right)$$

This results in slower overall emotional stabilization.

**Fear reduction during recovery**. Finally, the magnitude of fear reduction upon entering cooldown is also modulated. Instead of a fixed reduction, fear is scaled by a Neuroticism-dependent recovery factor:

$$F_i(t) \leftarrow F_i(t) \cdot r(\psi_i^N)$$

where $r(\psi_i^N) \in [0.3, 0.7]$ increases with Neuroticism. Consequently, low-Neuroticism agents experience stronger fear reduction and recover more effectively, while high-Neuroticism agents retain higher residual fear even during recovery.

## 3.6 Summary of Framework Extensions over ESCAPES

To clearly position the proposed framework with respect to the baseline ESCAPES framework, we summarize the main extensions introduced in this work. While improvements are discussed throughout the previous sections, consolidating them highlights the conceptual contributions and facilitates comparison. Overall, our framework extends ESCAPES along the following dimensions:

1. The representation of informational and emotional states is generalized from discrete to continuous domains. In ESCAPES, both Event Certainty Level and Fear Level are modeled as integer-valued variables with limited resolution. In contrast, the proposed modeling introduces continuous representations for both Event Certainty Level and Fear Level, enabling gradual belief updates and smoother behavioral transitions.
2. Unlike ESCAPES, where evacuation behavior is triggered implicitly through discrete *EventCertainty* levels and probabilistic behavior selection, the proposed framework introduces an explicit threshold-based decision mechanism. Each agent initiates evacuation only when its Event Certainty Level exceeds an individual-specific evacuation threshold. This mechanism captures heterogeneity in responsiveness to risk. By assigning heterogeneous thresholds across agents, the framework allows realistic variation in evacuation timing, reflecting empirical observations that individuals differ in how readily they act upon uncertain or incomplete information.
3. The process of knowledge acquisition is reformulated. While ESCAPES models knowledge spread through discrete updates driven by proximity and authority influence, the proposed framework introduces an event-driven, continuous accumulation mechanism. Social influence is modeled in a graded manner, where the impact of an interaction depends on both the certainty and the informational source of neighboring agents. This results in a more realistic representation of how information propagates under uncertainty.
4. Exit knowledge is significantly extended. In ESCAPES, exit knowledge is binary and static, with limited stochastic forgetting. In contrast, the proposed framework introduces a structured and dynamic representation of exit knowledge that separates acquisition from recall. It explicitly models temporary forgetting, time-dependent decay, and the influence of emotional state on memory retrieval. This allows the framework to capture cognitive degradation under stress, which is well-supported in empirical literature.
5. Beyond generic stochastic forgetting, the proposed framework introduces a parametric and interpretable formulation of forgetting probabilities that captures differences in spatial memory robustness across agent types. In particular, the base forgetting probability is defined as a function of agent characteristics (e.g, role, experience, or training level), enabling the representation of populations with varying levels of familiarity and cognitive stability. This allows the framework to distinguish, for example, between inexperienced agents with fragile memory representations and highly trained agents with stable recall. This provides a simple yet powerful mechanism to model the impact of training, experience, and environmental familiarity on evacuation performance – an aspect that is not explicitly captured in ESCAPES.

6. ESCAPES employs a simplified fear model that primarily influences movement speed and propagates via a maximum-value contagion rule. The proposed framework introduces a continuous and multi-component fear formulation, where fear arises from distinct sources, including direct exposure to danger, informational uncertainty, and stagnation during evacuation. These components are combined through a dominance mechanism, allowing the most critical stressor to govern emotional state. Additionally, panic is modeled as a high-intensity region within the fear spectrum rather than as a discrete category.
7. The role of fear is expanded beyond movement modulation. In the proposed framework, fear directly affects perception, memory, decision-making, and motor control. High fear levels degrade perceptual accuracy, impair recall of exits, and suppress rational planning, leading to reactive behavior.
8. Finally, unlike ESCAPES the proposed framework introduces personality-driven heterogeneity through the Neuroticism trait of the OCEAN model. Personality acts as modulation layer that influences fear amplification, emotional contagion, stagnation sensitivity, and recovery dynamics. This enables a unified and continuous representation of individual differences without relying on discrete behavioral categories.

Table 1: Key Differences and Justification

| **Aspect** | **ESCAPES** | **Proposed Framework** | **Improvement/Justification** |
|---|---|---|---|
| Event Awareness | Discrete Event Certainty {0,1,2} | Continuous ECL [0,1] | Captures gradual belief formation & uncertainty |
| Knowledge Spread | Rule-based, categorical | Graded, continuous influence | More realistic social information diffusion |
| Exit Knowledge | Binary, static | Structured, dynamic, with forgetting/Recall | Models stress-induced cognitive degradation |
| Memory | Simple stochastic forgetting | Fear and time-dependent forgetting | Aligns with cognitive psychology findings |
| Memory Robustness & Forgetting Heterogeneity | Not modeled (uniform stochastic) | Parametric, agent-specific forgetting probability (based on role/experience/training) | Enables modeling of training, experience, and cognitive stability differences across populations |
| Fear Representation | Discrete {0,1,2} | Continuous [0,1] | Enables smooth escalation and heterogeneity |
| Fear Sources | Implicit, aggregated | Multi-component (fire, uncertainty, stagnation) | Improves interpretability and realism |
| Emotional Contagion | Max neighbor fear | Gradual convergence (personality-based) | Avoids unrealistic abrupt transitions |
| Behavioral Impact of Fear | Mainly speed | Perception, memory, decision, motion | Stronger cognitive-behavior coupling |
| Panic Modeling | Not explicit | Emergent via threshold | Matches literature on panic as extreme fear |
| Decision Mechanism | BDI + probabilistic | Threshold-based + cognitive states | Simpler, more interpretable decision logic, enabling heterogeneity |
| Agent Heterogeneity | No Personality modeling | Personality (Neuroticism) | Aligns with literature that personality drives emotional responses during evacuation |
| Panic Recovery Dynamics | Not explicit | Panic cooldown mechanism | Models temporal emotions regulation |

### 3.7 Graph-Based Representation of the framework

To provide a unified view of the interactions between the different components of the proposed framework, we introduce a conceptual graph representation (Figure 1) that captures the causal dependencies between perception, cognition, decision, personality, emotion, and behavior.

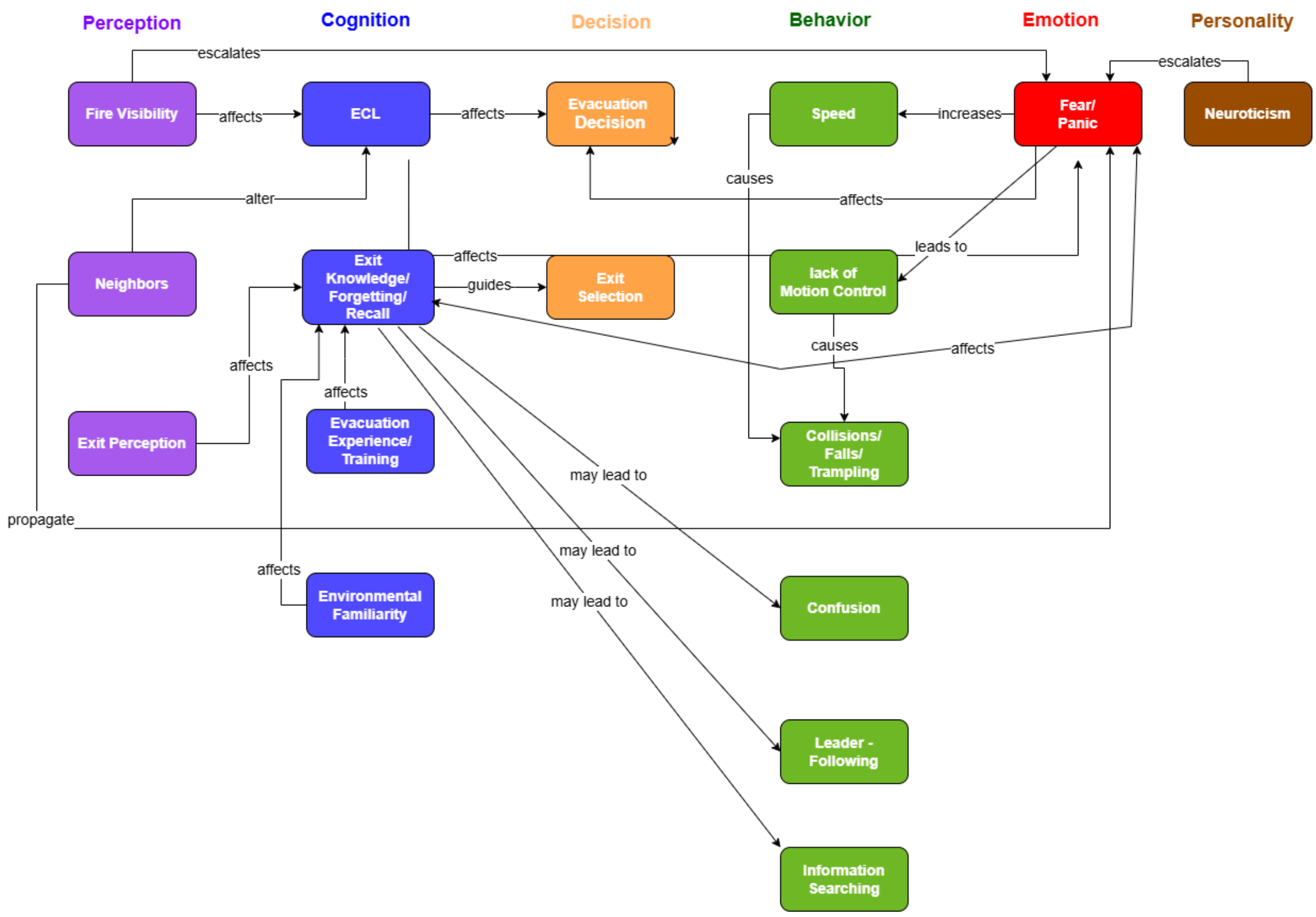

Figure 1: Model Graph

## 4 Experimental Methodology and Setup

### 4.1 Experimental Methodology

All experiments presented in this study follow a common experimental methodology designed to ensure controlled comparisons between different behavioral configurations of the agent population.

In each experiment, agents are organized into distinct groups representing different behavioral or cognitive configurations. Each group is characterized by a specific parameter setting (e.g. spatial familiarity, memory stability, evacuation threshold, while all remaining parameters remain identical. This controlled design ensures that any observed differences in evacuation outcomes can be attributed solely to the parameter under investigation.

For every agent group, 30 independent simulation runs are performed and each simulation run consists of a population of 30 agents. The building layout, environmental conditions, and event timing remain identical across runs, ensuring that all experiments are conducted under the same spatial and temporal conditions.

Although agents are initialized at predefined spatial locations, the simulation includes stochastic elements that introduce controlled variability between runs. In particular, agents perform random, non-goal-directed movement before the onset of the emergency event. As a result, the spatial configuration of agents at the moment the fire is triggered differs between runs, producing diverse evacuation dynamics while preserving identical initial conditions.

During each simulation, a set of behavioral and outcome metrics (defined in Section 4.3) is recorded at discrete logging intervals. These metrics include evacuation outcomes (*escaped, injured, dead, dead_after_injury*), state indicators (*fallen, trampled, following_leader, informed, panicking, inPanicCooldown, evacuating, confused, searchingForInformation*), and cognitive or emotional indicators (*avg_FearLevel, avg_EventCertaintyLevel, avg_knownExits*).

Finally, for each agent group, the values of the logged metrics are aggregated across the 30 simulation runs. Specifically, the mean value of each metric is computed at each time step and visualized in a time-series graph. In these graphs, the horizontal axis represents the simulation time step. The vertical axis represents either:

1. the proportion of agents exhibiting a specific state or outcome (e.g., escaped, injured, dead), expressed as a percentage of the total agent population,
2. the average value of a cognitive or internal variable across the living agents (e.g., average Fear Level, average Event Certainty Level, or average number of known exits).

The mean trajectory across runs is displayed as a line, while shaded regions indicate one standard deviation around the mean, illustrating the variability introduced by stochastic agent behavior across different simulation runs.

## 4.2 Simulation Environment and Scenario Setup

To assess how the proposed modeling of event awareness, exit knowledge, and fear influence evacuation outcomes such as evacuation success, injuries, and fatalities, we designed a controlled simulation environment. The environment represents the interior of a public building and provides a structured yet flexible spatial context for conducting repeatable evacuation experiments.

The building layout consists of two primary rooms (Room A and Room B), a central lobby area, and a corridor leading to the main exit of the building. In addition to the main exit, a secondary emergency exit is positioned in the lower-left region of the environment. Exit locations are visually represented within the scene as green cubes. A top-down view of the simulation environment is shown in Figure 2.

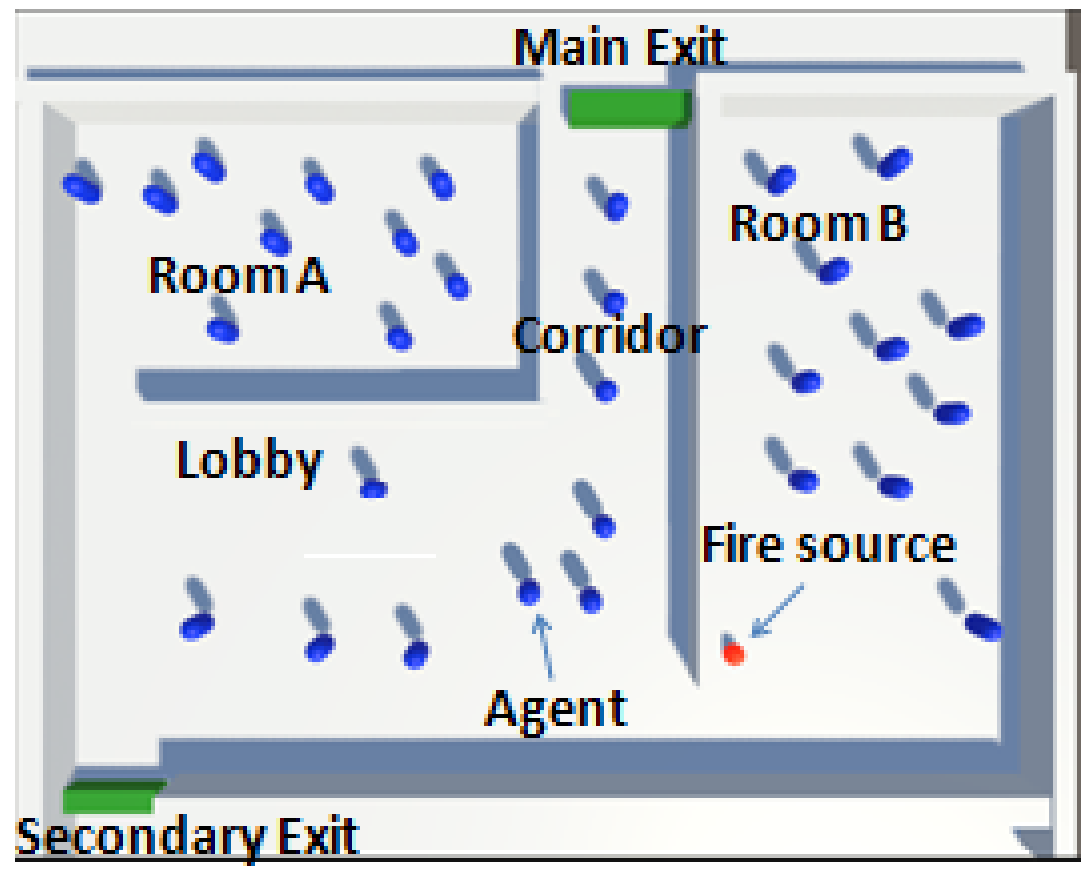

Figure 2: Top-down view of the simulation environment. The layout consists of two rooms (Room A and Room B), a central lobby, and a corridor leading to the main exit. A secondary exit is located in the lower-left region of the environment. Exit locations are indicated by green cubes, while agents are represented as blue cylinders at their initial predefined positions. Fire is represented as a red sphere located near the exit of Room B.

At initialization, agents – visually represented as blue cylinders – are placed at fixed, predefined positions within the interior spaces of the building as illustrated in Figure 1. The navigation of all agents are implemented through the A* algorithm, providing realistic navigation capabilities, including pathfinding, obstacle avoidance, and collision handling within the building layout. While RL-based navigation is also being investigated as part of ongoing work to support adaptive and experience-driven behaviors, A* is adopted in this framework to ensure stable, deterministic, and computationally efficient navigation, allowing the study to focus on the impact of cognitive and emotional processes on evacuation dynamics.

The emergency event is introduced after a predefined delay from the start of the simulation. This delay is controlled by a configurable parameter (*FireStartTime*), enabling precise synchronization across experimental runs. Once triggered, the fire is visualized as an expanding red sphere, representing a dynamically increasing hazardous region.

By maintaining a fixed spatial layout and controlled event timing, the simulation environment enables systematic experimentation under varying cognitive and emotional configurations. This design choice is intentional, as it allows us to isolate and systematically evaluate the effects of the proposed cognitive, emotional, and behavioral mechanisms without confounding influences from varying spatial configurations. The selected environment provides a sufficiently complex and representative setting – including multiple exits, spatial constraints, and limited visibility – to capture key evacuation dynamics, while ensuring experimental comparatibility across scenarios. While evaluating multiple environments constitutes an important direction for future work, the focus of this study is on validating the proposed framework components under controlled conditions rather than on environment-specific performance.

## 4.3 Definitions of Logged Metrics

Let:

$$A$$

be the set of all agents initialized in the simulation.

$$A_{alive}\,(t) \subseteq A$$

be the set of agents that are not dead at time $t$.

All metrics are evaluated at discrete logging steps:

$$t \;=\; k \cdot logInterval$$

a) **escaped**: The cumulative number of agents that have successfully reached an exit and left the hazardous environment (evacuated) by time step $t$.

$$escaped(t) = |\{a \in \mathrm{A} | a\ has\ exited\ the\ environment\ by\ t\}|$$

b) **injured**: The number of agents that are currently injured but still alive at time $t$.

$$injured(t) = |\{a \in \mathrm{A}_{alive}\,(t) | a\ has\ been\ injured\ by\ t\}|$$

c) **dead**: The cumulative number of agents that have died due to fire exposure, trampling, or accumulated injuries by time $t$.

$$dead(t) = |\{a \in \mathrm{A} | a\ is\ dead\ at\ t\}|$$

d) **dead_after_injury**: The cumulative number of agents that died after having previously been injured.

$$dead_after_injury(t) = |\{a \in \mathrm{A} | a\ is\ dead\ and\ was\ injured\ before\ death\ by\ t\}|$$

e) **fallen**: The number of agents that are currently in a fallen state at time $t$.

$$fallen(t) = |\{a \in \mathrm{A} | a.isFallen = true\ at\ t\}|$$

f) **trampled**: The number of agents that are currently being trampled by surrounding agents at time $t$.

$$trampled(t) = |\{a \in \mathrm{A} | a.isBeingTrampled = true\ at\ t\}|$$

Indicator of extreme crowd pressure and panic escalation.

g) **following_leader**: The number of agents that are currently following another agent rather than navigating independently.

$$following_leader(t) = |\{a \in \mathrm{A} | a.isFollowingLeader = true\ at\ t\}|$$

h) **informed**: The number of agents that are currently informed about the emergency event by an authority agent.

$$informed(t) = |\{a \in \mathrm{A} | a.isInformed = true\ at\ t\}|$$

i) **panicking**: The number of agents that are currently in a panic state (i.e., agents whose fear exceeds the panic threshold).

$$panicking(t) = |\{a \in \mathrm{A} | a.isPanicking() = true\ at\ t\}|$$

j) **inPanicCooldown**: The number of agents that are currently in a panic cooldown phase (agents temporarily immune to panic re-triggering).

$$inPanicCooldown(t) = |\{a \in \mathrm{A} | a.inPanicCooldown = true\ at\ t\}|$$

k) **evacuating**: The number of agents that have committed to evacuation behavior at time $t$.

$$evacuating(t) = |\{a \in \mathrm{A} | a.isEvacuating = true\ at\ t\}|$$

l) **confused**: The number of agents that are currently in a confused state due to lack of recalled exits.

$$confused(t) = |\{a \in \mathrm{A} | a.isConfused = true\ at\ t\}|$$

m) **searchingForInformation**: The number of agents that are actively searching for information at time $t$.

$$searching(t) = |\{a \in \mathrm{A} | a.isSearchingForInformation = true\ at\ t\}|$$

n) **avg_FearLevel**: The mean fear level among all living agents at time $t$.

$$avg_FearLevel(t) = \frac{1}{|\mathrm{A}_{alive}\ (t)|} \sum_{a \in \mathrm{A}_{alive}\ (t)} FearLevel_a(t)$$

where $FearLevel_a(t)$ is the Fear Level of agent $a$ at time $t$.

o) **avg_EventCertaintyLevel**: The mean event certainty level among all living agents at time $t$.

$$avg_EventCertaintyLevel(t) = \frac{1}{|\mathrm{A}_{alive}\ (t)|} \sum_{a \in \mathrm{A}_{alive}\ (t)} ECL_a(t)$$

where $ECL_a(t)$ is the event certainty level of agent $a$ at time $t$.

p) **avg_knownExits**: The average number of exits currently recalled per living agent at time $t$.

$$avg_knownExits(t) = \frac{1}{|A_{alive}(t)|} \sum_{a \in A_{alive}(t)} |\{e \in E_a | e.knows = 1 \wedge e.temporarilyForgotten = 0\}|$$

where $E_a$ denotes the exit knowledge set of agent a, consisting of one memory entry for each exit that the agent has encountered or initialized with during the simulation.

## 4.4 Experimental Design Summary

To facilitate comparison between the different experimental scenarios examined in this study, Table 2 summarizes the design of the six experiments conducted.

The experiments follow two complementary objectives. Experiment 1 serves as a baseline model comparison, evaluating the proposed cognitive – emotional agent framework against a classical evacuation model based on omniscient and purely rational agents. This experiment assesses the overall impact of incorporating event awareness, dynamic exit knowledge, and fear dynamics into evacuation modeling.

In contrast, Experiments 2-6 investigate the sensitivity of evacuation outcomes to specific behavioral parameters of the proposed framework. In these experiments, agents are organized into groups that differ only in one parameter at a time (e.g., spatial familiarity, memory robustness, evacuation decision threshold, activation of fear, or personality traits), while all other parameters remain identical. This controlled design ensures that the effect of each parameter on evacuation dynamics and safety outcomes to be examined independently.

Specifically, Experiments 2-4 focus on cognitive and decision-related factors, including spatial familiarity, memory robustness, and evacuation sensitivity. Experiment 5 complements the analysis by isolating the role of emotional dynamics, evaluating the impact of fear and panic on evacuation performance and the emergence of complex crowd behaviors.

Experiment 6 further extends this analysis by introducing personality into the framework, examining how individual differences in the Neuroticism trait modulate fear dynamics and influence both behavioral responses and safety outcomes. This experiment bridges the cognitive-emotional framework with stable personality characteristics, enabling the study of how intrinsic individual traits shape collective evacuation dynamics.

Across all experiments, the same simulation environment, agent population size, and experimental methodology are maintained, as described in Sections 4.1-4.3. This design choice is intentional, as it allows us to isolate and systematically evaluate the effects of the proposed cognitive, emotional, and behavioral mechanisms without confounding influences from other factors e.g. varying spatial configurations.

Table 2 provides an overview of the experimental objectives, the agent groups considered in each experiment, the parameter or modeling aspect under investigation, and the primary evacuation metrics analyzed.

Table 2: Overview of the experimental design and investigated behavioral factors

| Exp | Objective | Agent Groups | Modeling Aspect/ Parameter Investigated | Main Metrics Analyzed |
|---|---|---|---|---|
| 1 | Compare the proposed cognitive-emotional evacuation framework with a classical omniscient and purely rational agent model | OmniscientAndLogical,<br>NonOmniscientAndEmotional | Modeling comparison (classical rational agents vs. agents with event awareness, dynamic exit knowledge, and fear) | All metrics |
| 2 | Evaluate the impact of spatial familiarity on evacuation outcomes | FirstTimeVisitors, RegularVisitors, Staff | Initial knowledge of secondary exit and base forgetting probability | All metrics |
| 3 | Examine how staff experience influences exit recall and the | New Staff, Experienced Staff, Veteran Staff | Base forgetting probability | confused |
| | emergence of confusion | | | |
| 4 | Investigate how heterogeneity in evacuation decision sensitivity affects evacuation outcomes | Threshold = 0.2 (willing to evacuate),<br>Threshold = 0.5 (Neutral),<br>Threshold = 0.8 (unwilling) | Evacuation decision threshold | Escaped, injured, dead, dead_after-injury |
| 5 | Assess the impact of fear and panic on evacuation outcomes and emergent crowd behavior | Fear OFF, Fear ON | Activation of fear (FearEnabled: false vs. true) | Escaped, injured, dead_after_injury, fallen, trampled, following_leader,<br>searchingForInformation |
| 6 | Investigate the impact of personality (Neuroticism) on fear dynamics and evacuation outcomes | Neuroticism = 0.1 (low), 0.5 (medium), 0.9 (high) | Personality-driven modulation of fear generation, propagation, and recovery mechanism | avg_FearLevel, panicking, escaped, injured, dead, dead_after_injury, fallen, trampled |

# 5 Simulation Results

## 5.1 Experiment 1: Omniscient and Logical vs. Non-Omniscient and Emotional Agents

The first experiment evaluates the impact of the proposed modeling of event knowledge, exit knowledge, and fear on evacuation outcomes by comparing it against a classical omniscient and purely rational agent configuration.

Two agent groups are considered: *OmniscientAndLogical* and *NonOmniscientAndEmotional*. All agents are first-time visitors. Agents in the *OmniscientAndLogical* group represent an idealized evacuation model. They possess complete and persistent knowledge of all available exits and have immediate, perfect awareness of the emergency event (*EventCertaintyLevel* = 1 at fire onset). Emotional dynamics are disabled, resulting in fully rational, fear-free behavior throughout the evacuation.

Agents in the *NonOmniscientAndEmotional* group operate under the proposed models of event awareness, exit knowledge, and fear. Their knowledge of exits and the emergency evolves dynamically over time, and emotional responses influence both decision-making and movement behavior during evacuation.

The results of Experiment 1, presented in Figure 3, highlight the qualitative impact of incorporating dynamic event knowledge, exit knowledge, and fear into the evacuation model. The goal of this experiment is not a strict quantitative comparison, but a proof-of-concept demonstration that the proposed cognitive-emotional mechanisms produce more realistic evacuation dynamics.

In Figure 3a, the *OmniscientAndLogical* group reaches a maximum evacuation rate of 98% at t=37, whereas the *NonOmniscientAndEmotional* agents reach a maximum slightly above 30%. The fact that 100% evacuation is not achieved even in the omniscient case is explained by fire-induced fatalities. The substantial contrast between the two groups qualitatively confirms the expected influence of imperfect knowledge and emotional dynamics on evacuation efficiency. Importantly, the emphasis here is qualitative: different parameter values within the proposed models could lead to quantitatively different outcomes (e.g., higher or lower escape rates, injuries, or fatalities), but the observed direction of the effect aligns with theoretical expectations.

Across most logged safety and behavioral metrics, the *OmniscientAndLogical* group exhibits consistently zero values throughout the simulation: injuries (Figure 3b), fatalities (Figure 3c), fatalities after injuries (Figure 3d), falls (Figure 3e), trampling (Figure 3f), leader-following behavior (Figure 3g), panicking agents (Figure 3h), agents in panic cooldown (Figure 3i), confused agents (Figure 3k), agents searching for information (Figure 3l), and average fear level (Figure 3m). This reflects the deterministic and fully rational nature of the *OmniscientAndLogical* configuration. In contrast, the *NonOmniscientAndEmotional* agents exhibit non-zero values in these metrics, generating qualitatively realistic patterns.

Figure 3j further illustrates this contrast. In the *OmniscientAndLogical* model, 100% of agents decide to evacuate immediately at fire onset (t=6). Conversely, the *NonOmniscientAndEmotional* agents display a gradual increase in evacuation decisions over approximately 80 time steps, reproducing the well-documented pre-evacuation delay phenomenon.

A similar pattern appears in event awareness (Figure 3n). In the omniscient configuration, all agents become fully aware of the emergency at t=6, the exact moment of fire ignition. In the proposed framework, however, the average Event Certainty Level increases progressively over time, as agents either directly perceive the fire or acquire information through social interaction. This gradual awareness spread reflects more plausible emergency perception dynamics.

Finally, Figure 3o shows that *OmniscientAndLogical* agents possess complete and persistent knowledge of both exits from the start of the simulation (t=0) until its end (around t=30). In contrast, the *NonOmniscientAndEmotional* framework captures fluctuations in average exit knowledge over time, influenced by fear escalation, panic, confusion, information search, and forgetting.

Overall, the results demonstrate that incorporating dynamic cognition and emotion produces evacuation behavior that is qualitatively more realistic. While the exact quantitative outcomes depend on parameter calibration, the observed differences confirm that the proposed modeling framework meaningfully and coherently affects evacuation dynamics in the expected direction.

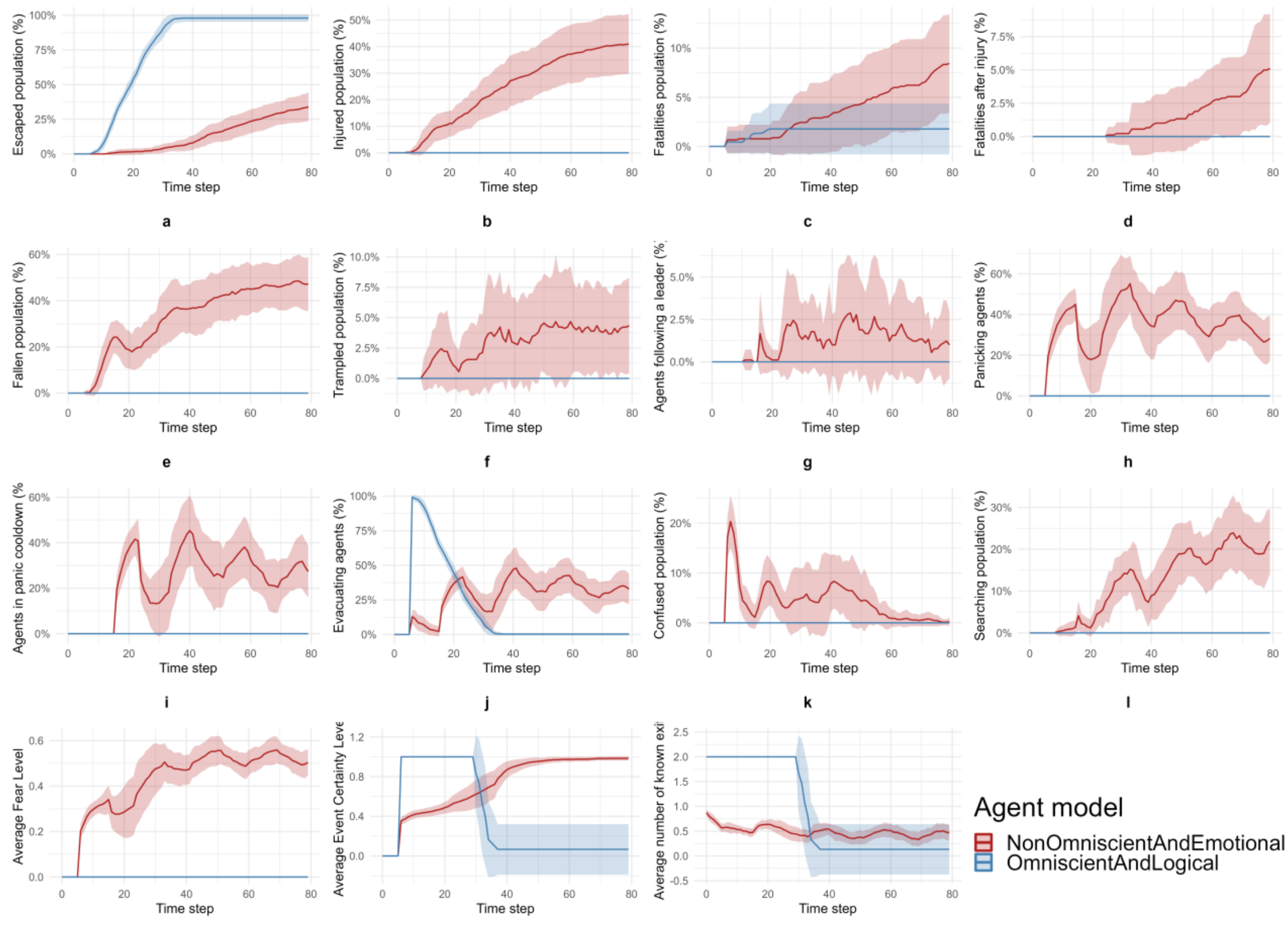

Figure 3: The outcomes of the 1st experiment

## 5.2 Experiment 2: Impact of Spatial Familiarity on evacuation outcomes

The second experiment investigates the impact of spatial familiarity on evacuation outcomes. In contrast to Experiment 1, which examined the role of cognitive and emotional modeling against an omniscient and logical baseline, Experiment 2 focuses exclusively on differences in prior environmental knowledge and the resulting memory stability across agent populations.

Three agent groups are considered: *FirstTimeVisitors*, *RegularVisitors*, and *Staff*. The groups differ in:

1. Their initial knowledge of the secondary exit, reflecting spatial familiarity, and
2. Their base forgetting probability $B(A_i)$, which governs the likelihood of losing previously acquired exit knowledge.

The *FirstTimeVisitors* group consists of agents who are all first time visitors to the environment. All agents are initialized with knowledge of the main exit, but none initially know the secondary exit. This group is assigned the highest base forgetting probability $B(A_i) = 0.3$, reflecting limited spatial familiarity and weaker memory robustness under stress.

The *RegularVisitors* group consists of agents who regularly visit the environment. All agents are initialized with knowledge of the main exit, while knowledge of the secondary exit is assigned probabilistically with a likelihood of 0.5. This group is characterized by a lower base forgetting probability $B(A_i) = 0.1$, representing increased familiarity and more stable memory retention compared to first-time visitors.

The Staff group consists of agents representing personnel working in the environment. All agents are initialized with knowledge of the main exit, and knowledge of the secondary exit is assigned with a high probability of 0.9. In addition, staff agents have a zero base forgetting probability $B(A_i)$, reflecting their previous training.

The objective of Experiment 2 is to evaluate how differences in spatial familiarity and memory stability influence evacuation efficiency and safety outcomes. By systematically varying only prior exposure and forgetting probability, the experiment isolates the contribution of environmental familiarity to overall evacuation performance.

The results of Experiment 2 presented in Figure 4, illustrate the qualitative impact of spatial familiarity on evacuation dynamics. As expected, increasing familiarity with the environment leads to improved evacuation performance and safer outcomes.

Figure 4a shows that the percentage of agents who successfully evacuate increases as spatial familiarity grows, with Staff achieving the highest escape rates, followed by *RegularVisitors*, and finally *FirstTimeVisitors*. Correspondingly, several adverse outcomes decrease with increasing familiarity. Lower familiarity groups exhibit higher percentages of injuries (Figure 4b), fatalities (Figure 4c), fatalities after injury (Figure 4d), fall incidents (Figure 4f), and trampling events (Figure 4g). Similarly, behavioral indicators associated with uncertainty and stress decrease as familiarity increases, including leader-following agents (Figure 4h), panicking agents (Figure 4j), agents in panic cooldown (Figure 4k), and information-seeking agents (Figure 4l). Overall, these patterns suggest that agents with greater knowledge of the environment navigate evacuation more efficiently and with fewer risky interactions.

Additional differences appear in agents' cognitive states. In Figure 4e, at the onset of the emergency, the percentage of confused agents exceeds 20% for *FirstTimeVisitors*, remains below 10% for *RegularVisitors*, and is zero for Staff. This result follows directly from the modeling definition: agents enter the confusion state when they are unable to recall any exit. Since staff agents have zero forgetting probability and all begin with knowledge of at least the main exit, confusion does not occur in this group.

Decision-making patterns are shown in Figure 4i. At the beginning of the emergency, higher familiarity leads to a larger proportion of agents immediately deciding to evacuate, with Staff agents exhibiting the highest initial evacuation decision rate. As the simulation progresses, this pattern gradually reverses. The explanation is that staff agents evacuate earlier, leaving fewer agents remaining in the environment to transition into the evacuating state at later timesteps.

Exit knowledge dynamics are presented in Figure 4o. As expected, higher spatial familiarity corresponds to a greater average number of known exits. However, *FirstTimeVisitors* and *RegularVisitors* display a gradual decline in this metric over time due to the effects of exit forgetting. In contrast, Staff agents remain consistently close to an average of two known exits, since they are unaffected by forgetting and the vast majority are initialized with knowledge of both exits.

Figure 4m presents an interesting effect regarding average Event Certainty Level. Surprisingly, staff agents exhibit lower certainty values throughout the simulation. This occurs because staff agents tend to decide to evacuate earlier, as shown in Figure 4i. Consequently, agents located in the room where the fire originates quickly move toward exits instead of interacting with other agents. As a result, social transmission of information about the event becomes weaker, delaying the spread of knowledge to agents located in the lobby and in other rooms. These agents primarily become aware of the fire later, when it grows large enough to be directly visible. Thus, increasing familiarity indirectly reduces the spread of event information between agents, leading to a lower average certainty level.

Finally, Figure 4n does not reveal a clear or systematic difference in average fear levels among the three familiarity groups.

Overall, the results demonstrate that increasing spatial familiarity improves evacuation efficiency and reduces hazardous interactions, confirming the qualitative influence of prior environmental knowledge on crowd evacuation dynamics. These findings are consistent with empirical observations indicating that individuals familiar with an environment tend to respond more proactively during emergencies, quickly identifying the nearest visible exit and proceeding without waiting for instructions [69].

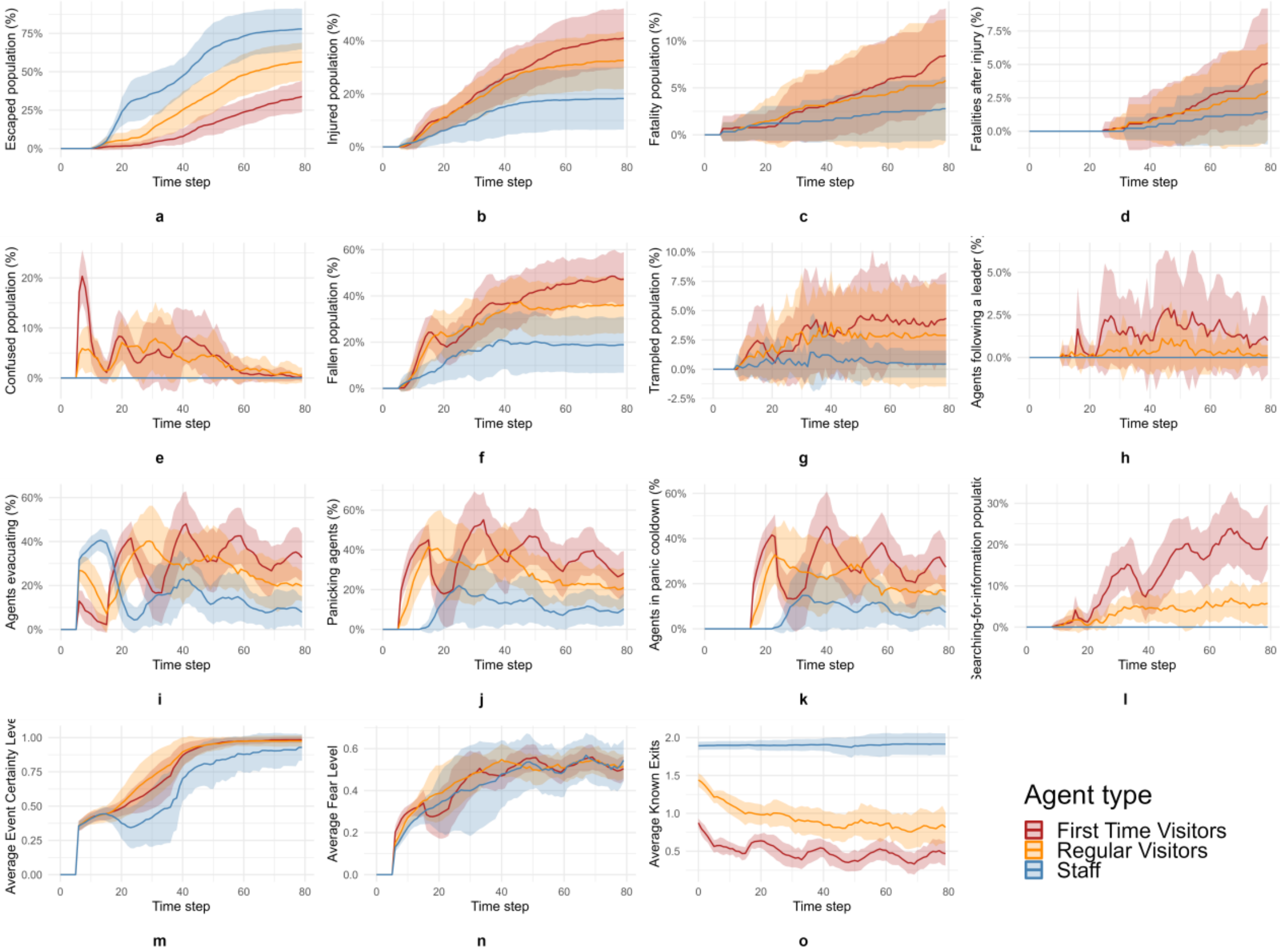

Figure 4: The outcomes of the 2nd experiment

## 5.3 Experiment 3: Impact of Staff Experience on Exit Recall and Confusion

The third experiment examines the impact of staff experience and training on exit recall and the emergence of confusion during evacuation. This experiment extends the analysis of spatial familiarity presented in Experiment 2 by focusing specifically on differences in memory robustness among staff members with varying levels of experience in the environment.

Three agent groups are considered: New Staff, Experienced Staff, and Veteran Staff.

All staff agents are initialized with knowledge of both exits. This is implemented through a Secondary Exit Knowledge Probability equal to 1.0, reflecting the common practice of informing employees about the available building exits during recruitment process and initial training. Consequently, all agents begin the simulation with complete exit knowledge. The groups differ only in their level of seniority, which is modeled through different forgetting probabilities. New Staff agents represent recently employed personnel with limited exposure to the environment and are assigned a higher forgetting probability $B(A_i) = 0.1$. Experienced Staff agents represent personnel with several years of experience in the environment and are assigned a lower forgetting probability $B(A_i) = 0.05$. Veteran Staff agents represent highly experienced personnel and are assigned a zero forgetting probability $B(A_i) = 0$, reflecting stable and persistent spatial memory.

In contrast to the previous experiments, this experiment focuses exclusively on the confusion state, which in the proposed model occurs when an agent is unable to recall any known exit. By isolating this metric, the experiment directly evaluates the relationship between spatial memory robustness and the likelihood of temporary cognitive paralysis under emergency conditions.

The objective of this experiment is to demonstrate that the proposed modeling framework can capture the effect of staff experience on evacuation behavior through a simple parameter adjustment. By varying the forgetting probability across otherwise identical staff populations, the model enables the emergence of different levels of confusion during evacuation, reflecting differences in spatial memory stability associated with varying levels of experience.

The results of Experiment 3 are presented in Figure 5a, which illustrates the temporal evolution of the proportion of agents in the confusion state for the three staff experience levels. A clear difference emerges between the groups. The New Staff group exhibits the highest proportion of confused agents mainly during the early stages of the evacuation. In contrast, the Experienced Staff group shows consistently lower confusion levels throughout the simulations. Although some temporary confusion still occurs, the overall magnitude of the effect is reduced compared to the New Staff group, reflecting the increased robustness of spatial memory associated with longer exposure to the environment. Finally, the Veteran Staff group exhibits no confusion events. Since these agents are assigned a zero forgetting probability, they retain stable knowledge of the exits even under stress conditions and therefore never enter the confusion state. As a result, the confusion curve for this group remains at zero for the entire stimulation duration.

Overall, the results demonstrate a monotonic relationship between staff experience and the likelihood of confusion. As staff experience increases, the proportion of agents experiencing temporary cognitive paralysis decreases. This behavior emerges directly from the memory dynamics of the framework, where more experienced agents are less susceptible to stress-induced memory degradation. The experiment therefore confirms that the proposed modeling framework can present differences in staff experience and training through variations in forgetting probability, allowing the model to capture their impact on evacuation outcomes (Figure 5b-c). These findings are consistent with previous observations that failure to act – manifested as immobility of 'freezing' – constitutes an impaired behavioral response that delays evacuation and may create a self-reinforcing process leading to fatalities even in otherwise survivable conditions [46].

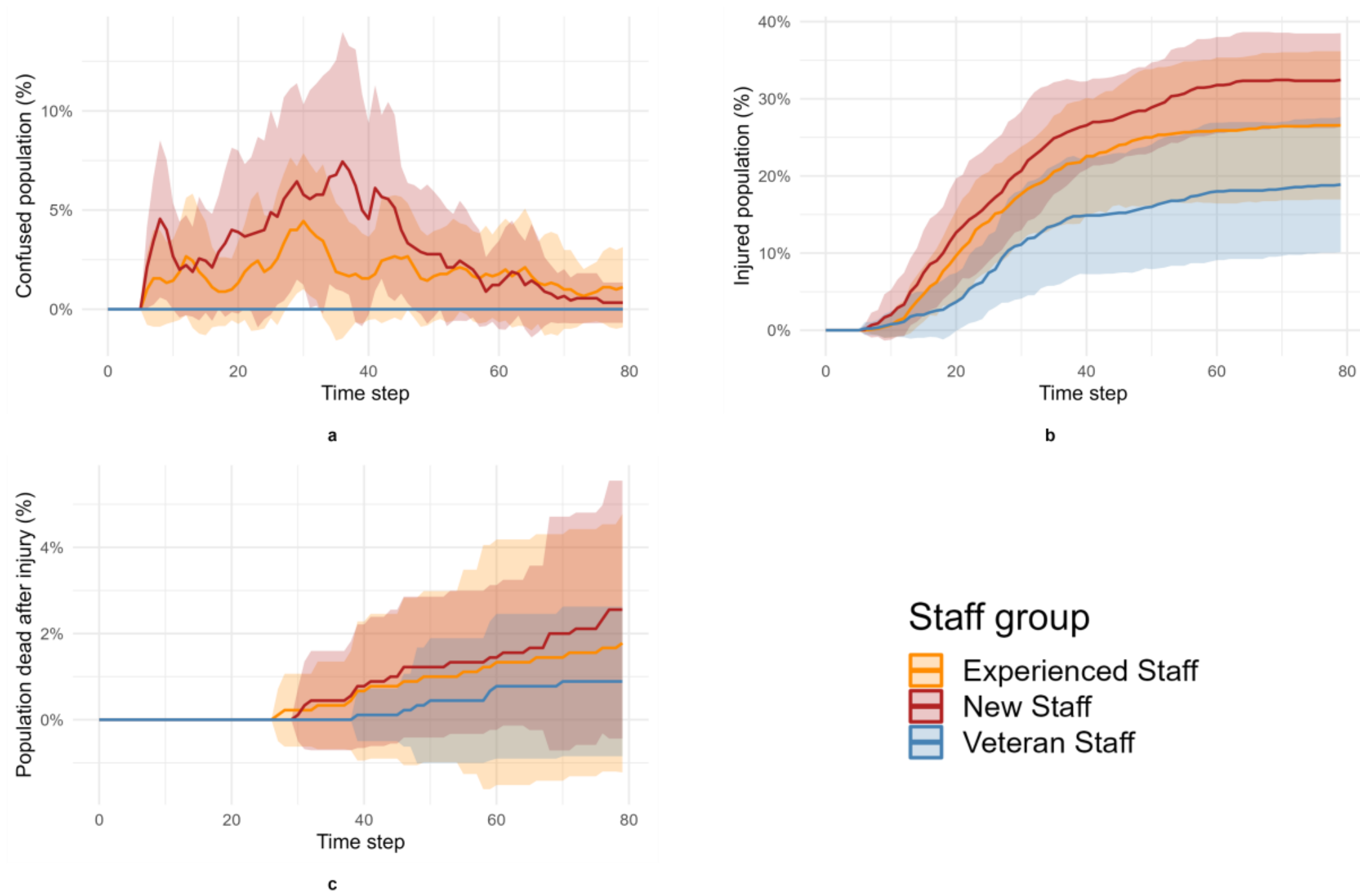

Figure 5: The outcomes of the 3rd experiment

## 5.4 Experiment 4: Impact of Evacuation Threshold Heterogeneity on Evacuation Outcomes

The fourth experiment investigates how heterogeneity in evacuation decision thresholds influences evacuation dynamics and outcomes. In the proposed framework, evacuation decisions follow a threshold-based mechanism: during each update cycle, agents evaluate whether their Event Certainty Level exceeds their personal evacuation threshold. Only when this condition is satisfied does an agent commit to evacuation behavior. By assigning this threshold individually to agents, the framework captures population heterogeneity and allows the emergence of realistic behavioral patterns such as hesitation, delayed reactions, and responses to uncertain or incomplete information.

This mechanism also provides a natural way to connect evacuation behavior with personality characteristics of individuals. In particular, the evacuation threshold can represent individual differences in how readily individuals react to perceived danger [54]. Some individuals may react quickly even under limited certainty, while others may require stronger confirmation before deciding to evacuate, because they have prior experience with physical disasters [54], or due to a rich accident experience [8].

To explore this effect, three agent groups are defined, representing different behavioral tendencies:

- Agents willing to evacuate with a low evacuation threshold (0.2), who tend to react quickly to emerging threats.
- Neutral agents with a medium evacuation threshold (0.5), representing typical evacuation behavior.
- Unwilling agents with a high evacuation threshold (0.8), who require strong confirmation of the event before committing to evacuation.

Three corresponding agent groups are therefore considered: Threshold= 0.2, Threshold = 0.5, and Threshold = 0.8. All agents are staff agents.

The results of Experiment 4 are presented in Figure 6. As unwillingness increases, and consequently the evacuation threshold becomes higher, fewer agents initiate evacuation decisions following the onset of the fire (Figure 6a). This delayed evacuation response leads to lower escape rates over time (Figure 6b), and poorer safety outcomes, including increased injury rates (Figure 6c), higher fatality rates (Figure 6d), and a greater rate of fatalities occurring after injury (Figure 6e).

By systematically varying only the evacuation threshold, the experiment demonstrates how the proposed modeling framework can capture behavioral diversity associated with personality differences and how these differences influence evacuation outcomes.

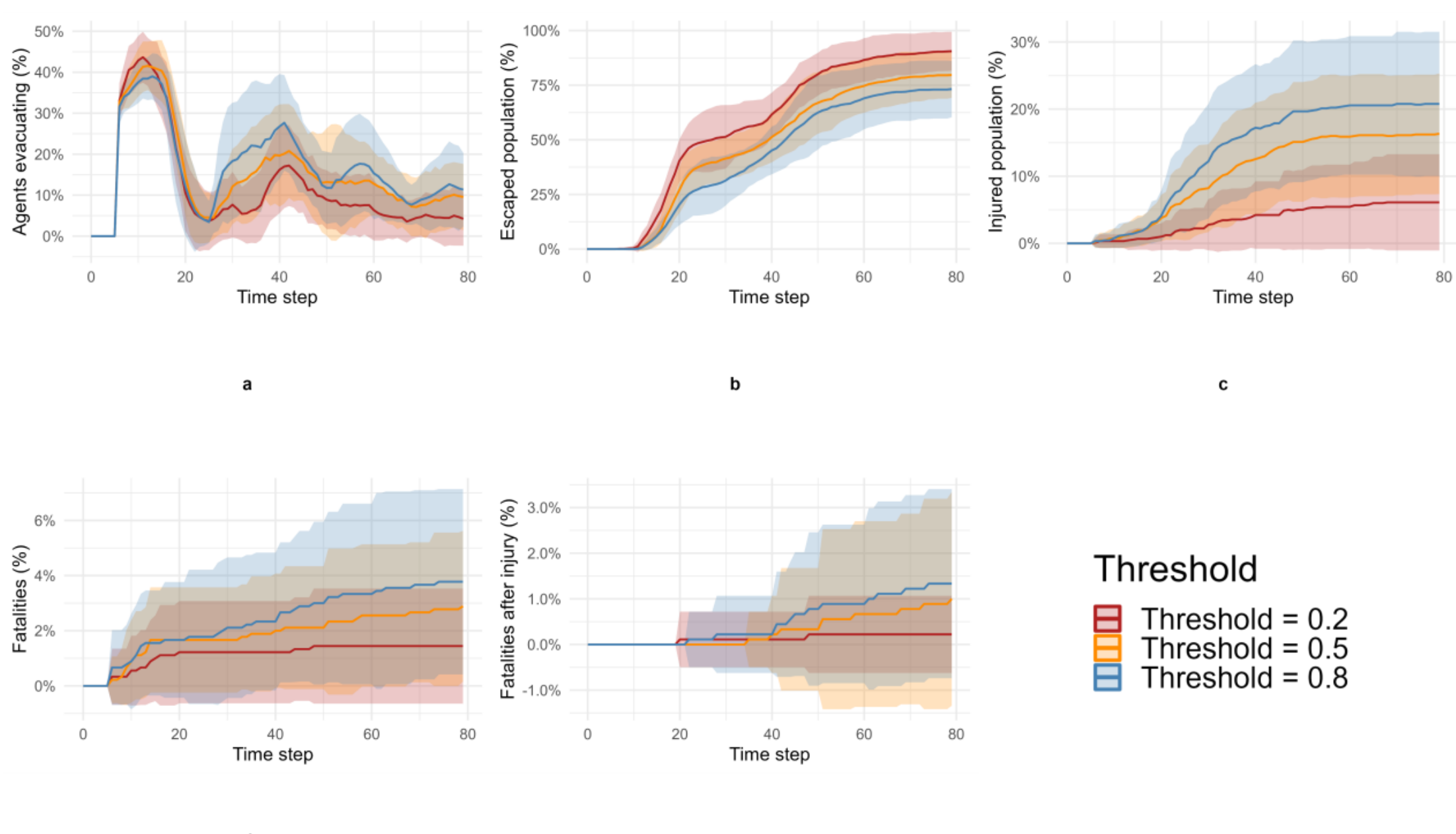

Figure 6: The outcomes of the 4th experiment

## 5.5 Experiment 5: Impact of Fear and Panic on Evacuation Outcomes

The fifth experiment investigates the impact of fear and panic on evacuation outcomes, with the objective of isolating the contribution of the proposed emotional modeling component. In contrast to previous experiments, which examined cognitive and behavioral parameters, this experiment focuses exclusively on the role of fear dynamics in shaping evacuation behavior and safety outcomes.

Two agent groups are considered: *Fear OFF* and *Fear ON*. In both groups, all agents are *FirstTimeVisitors* and share an identical evacuation threshold ($\theta = 0.5$). The two groups differ only in the activation of the fear model. In the *Fear OFF* group, emotional dynamics are disabled ($FearEnabled = false$), and agents operate without fear or panic. In the *Fear ON* group, emotional dynamics are enabled ($FearEnabled = true$), and agents experience fear and panic according to the proposed model. This controlled design ensures that any observed differences can be attributed solely to the presence of fear.

The results, presented in Figure 7, demonstrate a substantial impact of fear on both evacuation efficiency and safety outcomes. Figure 7a shows that the *Fear OFF* group achieves significantly higher evacuation rates. This indicates that fear and panic reduce overall evacuation efficiency.

A consistent pattern emerges across all safety-related metrics. The *Fear OFF* group exhibits zero injuries, zero fatalities after injury, and no occurrences of falls or trampling throughout the simulation. In contrast, the *Fear ON* group displays steadily increasing levels of injuries, fatalities after injury, falls, and trampling following the onset of the fire. These results indicate that harmful physical phenomena arise exclusively in the presence of fear and panic.

Behavioral differences between the two groups further support this observation. As shown in Figure 7f-g, agents in the *Fear ON* group exhibit higher levels of leader-following behavior and increased engagement in information-seeking actions. These patterns reflect elevated uncertainty and reduced decision autonomy under emotional stress. In contrast, *Fear OFF* agents display more stable and independent behavior, resulting in more efficient evacuation trajectories.

Taken together, the results demonstrate that fear acts as a critical driver of emergent evacuation dynamics. Rather than simply influencing individual decision-making, the presence of fear gives rise to complex system-level phenomena, including congestion, physical instability, and cascading failures that lead to injuries and fatalities. This confirms that the proposed modeling of fear is essential for capturing realistic evacuation behavior, as it directly generates both behavioral adaptations and adverse safety outcomes observed in real-world emergency scenarios.

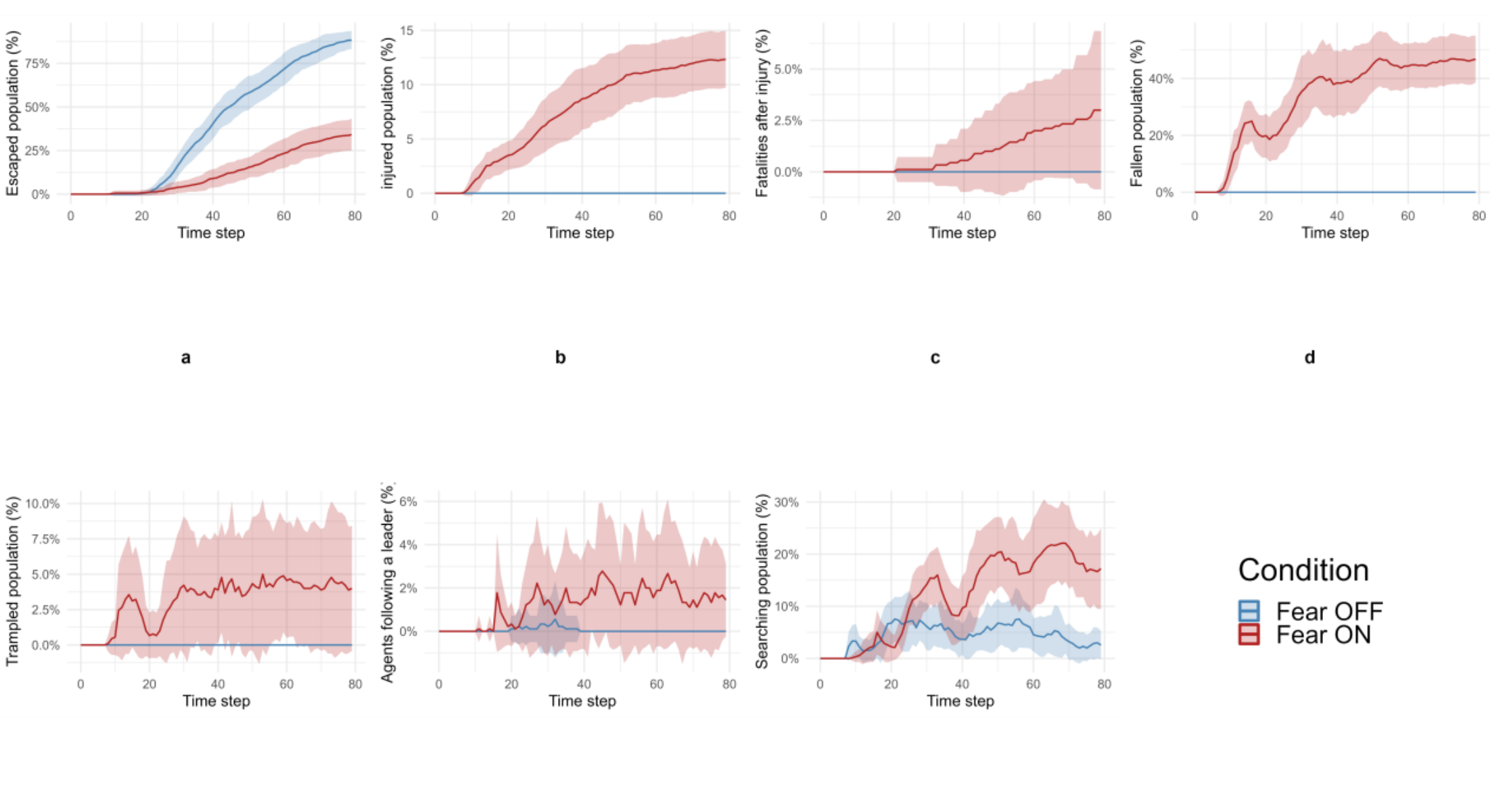

Figure 7: The outcomes of the 5th experiment

## 5.6 Experiment 6: Impact of Neuroticism on Fear and Evacuation Dynamics

The sixth experiment investigates the role of personality-specifically the Neuroticism trait – in shaping emotional responses and evacuation outcomes. While Experiment 5 established the importance of fear as a driver of evacuation dynamics, this experiment examines how individual differences in personality modulate that fear and consequently alter system-level behavior.

Three agent groups are considered, corresponding to different fixed values of the Neuroticism parameter:

- Low Neuroticism ($\varphi.N = 0.1$)
- Medium Neuroticism ($\varphi.N = 0.5$)
- High Neuroticism ($\varphi.N = 0.9$)

All other parameters of the framework remain identical across groups, including evacuation threshold ($\theta = 0.5$), agent type (*FirstTimeVisitors*), and all cognitive and environmental settings. This controlled design ensures that any observed differences are solely attributable to the modulation of fear by the Neuroticism trait, as introduced in the extended fear modeling.

The results, presented in Figure 8, reveal a clear and consistent effect of Neuroticism on both emotional dynamics and evacuation outcomes.

Figure 8a shows the evolution of the average fear level over time. A monotonic relationship is observed, with higher Neuroticism leading to systematically higher fear levels throughout the simulation. This confirms that the proposed modulation mechanism successfully amplifies fear responses for emotionally unstable agents.

This effect propagates directly to panic behavior. As illustrated in Figure 8b, higher-Neuroticism populations transition more rapidly into panic states, indicating increased sensitivity to perceived threat.

The escalation of panic has significant consequences for crowd safety. Figure 8c-d demonstrates that trampling and falling incidents increase substantially with higher Neuroticism levels. These phenomena emerge earlier and reach higher peaks in the high-Neuroticism group, reflecting increased physical instability and crowd pressure under intensified emotional conditions.

Evacuation performance follows an inverse trend. As shown in Figure 8e, lower Neuroticism leads to higher evacuation rates, with low-Neuroticism agents achieving the greatest proportion of escaped individuals. In contrast, higher Neuroticism reduces evacuation efficiency, as elevated fear and panic disrupt coordinated movement and decision-making.

Safety-related metrics reveal a clear trade-off. Figure 8f-h indicates that injuries, fatalities, and fatalities following injury, all increase with higher Neuroticism. The high-Neuroticism group consistently exhibits the worst safety outcomes; with more agents being injured and dying.

Overall, the results demonstrate that Neuroticism acts as a key modulator of emotional intensity, amplifying fear and accelerating panic propagation. This amplification produces stronger and earlier emergent phenomena, including trampling and physical instability. Although higher Neuroticism can increase evacuation speed, it significantly degrades safety, leading to more severe outcomes. These findings highlight the importance of incorporating personality traits into evacuation modeling. By linking stable individual characteristics to emotional responses, the framework captures a critical layer of human heterogeneity that directly influences both individual behavior and collective dynamics.

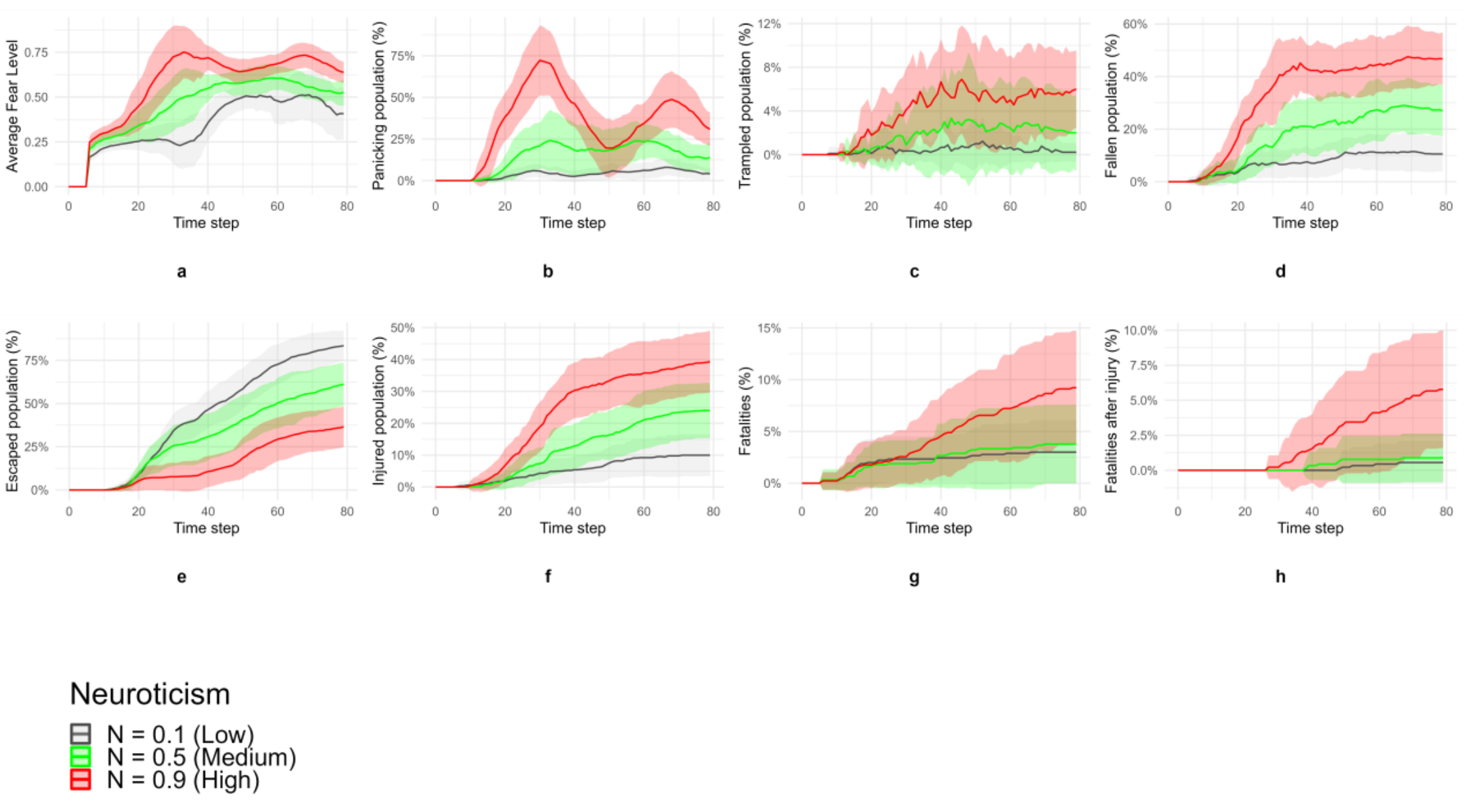

Figure 8: The outcomes of the 6th experiment

# 6 Discussion

The proposed agent-based evacuation framework extends traditional crowd evacuation simulations by incorporating cognitive, emotional, and personality-driven mechanisms that influence individual behavior during emergencies. While many existing models focus primarily on physical processes such as path planning, collision avoidance, and movement dynamics, previous research has emphasized the critical role of psychological factors – including perception, emotion, decision-making, and personality - in shaping evacuation behavior [67]. In particular, behavioral simulations have often emphasized physiological interactions while giving comparatively less attention to psychological variables such as emotion and personality traits [43].

The present framework addresses this gap by introducing mechanisms for uncertainty, social influence, emotional dynamics, and memory-based exit knowledge, and further extending them through personality-aware modulation. The introduction of a continuous Event Certainty Level allows agents to accumulate evidence about the existence of an emergency over time, rather than reacting instantaneously. This supports delayed responses, heterogeneous awareness, and asynchronous evacuation decisions. Additionally, individualized evacuation thresholds capture differences in agents' willingness to act, reflecting empirical findings that evacuation decisions depend on risk perception, prior experience, and information processing [67].

Emotional dynamics are explicitly modeled through fear and panic states, which influence both decision-making and movement. Agents may exhibit delayed responses or freezing behavior depending on their emotional state, consistent with research highlighting the interaction between psychological factors and behavioral responses in emergencies [8]. In this framework, emotion, cognition, and social interaction form a coupled system, where each component dynamically affects the others.

A key extension of the framework is the integration of personality, specifically the Neuroticism trait, as a systematic modulator of emotional dynamics. Rather than introducing new behavioral rules or predefined behavior categories, personality operates at a deeper level by shaping how agents perceive, amplify, and propagate fear. Neuroticism influences multiple stages of the fear process, including the intensity of fire-induced and uncertainty-driven fear, the rate of stagnation-induced escalation, susceptibility to emotional contagion, and recovery dynamics following panic states. This allows the framework to capture stable individual differences that lead to heterogeneous emotional trajectories even under identical environmental conditions.

The representation of exit knowledge as a dynamic cognitive process subject to acquisition, forgetting, and recall remains another key contribution. Unlike models that assume perfect or static knowledge, this approach reflects empirical observations that individuals may fail to recall known exits or prefer familiar routes under stress. The use of cognitive salience weights further captures familiarity-driven navigation, aligning with findings that occupants often exit through known paths rather than optimally distributed routes [67].

The experimental results further support the importance of integrating psychological and personality-related factors. The comparison between fear-enabled and fear-disabled agents demonstrates that emotional dynamics significantly affect evacuation outcomes, leading to reduced efficiency and the emergence of complex crowd phenomena such as injuries, falls, trampling, and increased reliance on social behaviors. Extending this analysis, the inclusion of personality reveals that these effects are not uniform across agents. Higher levels of Neuroticism systematically amplify fear, accelerate the onset of panic, and intensify its propagation, resulting in earlier and more severe emergent phenomena. At the same time, lower Neuroticism is associated with more stable emotional responses and improved evacuation efficiency.

These findings highlight that evacuation dynamics are not driven solely by external conditions or cognitive processes, but also by intrinsic individual characteristics. By linking stable personality traits to emotional responses, the framework captures an additional layer of heterogeneity that directly affects both individual decision-making and collective system behavior.

Despite these contributions, the proposed framework does not capture the full spectrum of human behavior observed in real emergencies. Empirical studies show that individuals exhibit diverse responses, including calm movement, observation, waiting, and protective actions, rather than uniform panic-driven behavior [43]. Moreover, cooperative and supportive behaviors frequently emerge depending on contextual and social factors [69].

While the framework reproduces several emergent behaviors – such as freezing, information seeking, and leader-following – it does not explicitly incorporate pro-social interactions, such as helping unfamiliar individuals or coordinated group behavior. Additionally, although personality is partially introduced through the Neuroticism trait, the remaining dimensions of the OCEAN model are not yet integrated, limiting the breadth of behavioral diversity that can be represented. Given the documented importance of both social cooperation and multidimensional personality effects in real evacuations, these represent key directions for future work.

More broadly, realistic evacuation modeling remains inherently challenging due to the complexity of human behavior and the absence of a universally accepted set of behavioral variables. As noted in prior work, models typically include only a subset of relevant factors due to conceptual and computational constraints [67]. Consequently, evacuation modeling should be viewed as an incremental process, where additional cognitive, emotional, social, and personality-related mechanisms are progressively integrated as empirical knowledge and computational capabilities advance.

# 7 Conclusions and Future Work

This paper presented an agent-based evacuation framework that integrates cognitive, emotional, and personality-driven mechanisms to simulate more realistic human behavior during emergency situations. In contrast to traditional evacuation models that assume perfect knowledge and fully rational agents, the proposed framework incorporates dynamic event awareness, memory-based exit knowledge, and a continuous fear-panic model that directly influences perception, decision-making, and movement behavior. This framework is further extended through the integration of personality, enabling the representation of stable individual differences in emotional response.

The simulation results demonstrate that these mechanisms significantly affect evacuation dynamics and outcomes. The comparison with an omniscient and purely rational baseline (Experiment 1) highlights the importance of modeling imperfect knowledge and emotional processes, as their inclusion leads to delayed evacuation decisions, heterogeneous awareness, and the emergence or realistic behavioral patterns. Experiments 2 and 3 further show that spatial familiarity and memory robustness play a critical role in evacuation efficiency, influencing confusion levels, exit recall, and overall safety outcomes. Experiment 4 demonstrates that heterogeneity in evacuation decision thresholds leads to substantial differences in response timing and evacuation success, emphasizing the importance of representing individual variability in risk perception and decision-making.

Experiment 5 provides a proof-of-concept validation of the proposed fear-panic modeling. The results indicate that the inclusion of fear dynamics reduces evacuation efficiency and gives rise to a range of emergent phenomena, including injuries, falls, trampling, and fatalities following injury. In contrast, simulations without fear produce unrealistically safe and orderly evacuations, with no physical interactions or adverse outcomes. These findings confirm that emotional dynamics are not merely descriptive variables, but fundamental drivers of both individual behavior and collective crowd phenomena.

Experiment 6 further demonstrates that these emotional dynamics are strongly modulated by personality. Specifically, the Neuroticism trait systematically amplifies fear, accelerates the onset and propagation of panic, and intensifies its impact on both behavior and safety outcomes. Higher Neuroticism leads to earlier and more severe emergent phenomena, including increased falls, trampling, injuries, and fatalities, while lower Neuroticism is associated with more stable emotional responses and improved evacuation performance. These results highlight that evacuation dynamics are shaped not only by environmental conditions and cognitive processes, but also by intrinsic individual characteristics.

Overall, the proposed framework demonstrates that integrating cognitive, emotional, and personality-related processes into agent-based simulations enhances behavioral realism and enables the emergence of complex evacuation patterns that cannot be captured by purely physical or deterministic approaches. By linking perception, cognition, emotion, and personality within a unified framework, the framework captures multiple layers of human behavior that jointly influence evacuation dynamics. While the quantitative outcomes depend on parameter calibration, the consistent qualitative trends observed across experiments support the validity and usefulness of the modeling framework.

Several directions for future research arise from this work. A key extension concerns the incorporation of pro-social behaviors, such as helping, coordination, and affiliation with familiar individuals, which are frequently observed in real evacuations but are not explicitly modeled in the current framework. Another important direction involves extending the personality modeling beyond Neuroticism to incorporate the remaining dimensions of the OCEAN model, enabling richer and more diverse behavioral profiles. This would allow the investigation of how different personality traits interact with cognitive and emotional processes to shape evacuation dynamics.

The emotional model itself could be further expanded to include additional affective states beyond fear and panic, allowing for more nuanced behavioral responses. Further improvements may also include the integration of additional human factors, such as heterogeneous physical characteristics (e.g., mobility limitations, body dimensions, or carried items), which can influence movement efficiency and congestion patterns. Finally, future work will focus on the empirical validation and calibration of the framework using real-world evacuation data, including video-based observations, in order to strengthen its predictive capability and practical applicability. As computational resources and data availability continue to improve, such extensions will be essential for advancing the realism and reliability of evacuation simulations.